\documentclass[lettersize,journal]{IEEEtran}
\usepackage{amsmath,amsfonts}
\usepackage{array}
\usepackage[caption=false,font=normalsize,labelfont=sf,textfont=sf]{subfig}
\usepackage{textcomp}
\usepackage{stfloats}
\usepackage{url}
\usepackage{verbatim}
\usepackage{graphicx}
\usepackage{cite}
\usepackage{booktabs}
\usepackage{pdfpages}
\usepackage{ulem}
\usepackage{color}

\usepackage{algorithm}
\usepackage{algpseudocode}%
\usepackage{siunitx}

\usepackage{booktabs}
\usepackage{amssymb}
\usepackage{amsmath} 
\usepackage{multirow}
\usepackage{pifont}
\usepackage{utfsym}

\usepackage{hyperref} 
\usepackage{float}
\usepackage{bm}
\usepackage[table]{xcolor}
\usepackage{colortbl}
\usepackage{dsfont} 
\definecolor{darkgreen}{RGB}{0,75,0}

\hypersetup{
    colorlinks=true,      
    linkcolor=red,         
    citecolor=green,       
    urlcolor=black,         
    linkbordercolor=white, 
    citebordercolor=white, 
    urlbordercolor=white   
}

\begin{document}

\title{Hierarchical Consistency Learning for Test-time Adaptation in Camouflage Perception}

\author{Mingfeng Zha, Tianyu Li, Guoqing Wang,~\IEEEmembership{Member,~IEEE,} Yunqiang Pei, Chaofan Qiao, Jiening Zhang\\ Yang Yang,~\IEEEmembership{Senior Member,~IEEE,} and Heng Tao Shen,~\IEEEmembership{Fellow,~IEEE}

\thanks{This work was supported in part by the National Natural Science Foundation of China under grant U23B2011, 62102069, U20B2063 and 62220106008, the Key R\&D Program of Zhejiang under grant 2024SSYS0091, the Sichuan Science and Technology Program under Grant 2024NSFTD0034.}

\thanks{Mingfeng Zha, Tianyu Li, Guoqing Wang, Yunqiang Pei, Chaofan Qiao, Jiening Zhang and Yang Yang are with the Center for Future Media and School of Computer Science and Engineering, University of Electronic Science and Technology of China, Chengdu 611731, China. (Email: zhamf1116@gmail.com)} 

\thanks{Heng Tao Shen is with the School of Computer Science and Technology, Tongji University, Shanghai 201804, China, with the Center for Future Media and School of Computer Science and Engineering, University of Electronic Science and Technology of China, Chengdu 611731, China, and with the Peng Cheng Laboratory, Shenzhen 518066, China.}

\thanks{Corresponding author: Tianyu Li.}

\thanks{Project page: https://winter-flow.github.io/project/HCL}

% \thanks{Manuscript received June 24, 2025; revised June 24, 2025.}

}

% \markboth{IEEE TRANSACTIONS ON IMAGE PROCESSING}%

\makeatletter
\def\ps@IEEEtitlepagestyle{
  \def\@evenfoot{}
}

\maketitle

\begin{abstract}
Camouflaged object detection (COD) aims to localize targets that exhibit minimal perceptual differences from backgrounds through physical attributes. Existing methods, constrained by the static train-then-freeze paradigm, suffer from domain rigidity and annotation dependency, limiting their adaptability to scene variations and unseen camouflage patterns. To overcome these, we propose the hierarchical consistency learning (HCL) framework, which integrates test-time adaptation for dynamic representation recalibration. Specifically, we design the hierarchical representation reconstruction (HRR) to alleviate feature entanglement by synergizing spatial reconstruction with dual-stream frequency-domain decomposition, enhancing robustness against appearance homogenization. The pixel and spectrum inference provide structural and contextual priors. We further introduce task affinity guidance (TAG) to propagate knowledge across branches via channel-wise affinity, aligning local discriminative cues and mitigating semantic drift. To ensure semantic invariance, we formulate the prototype consistency calibration (PCC), which aggregates region features into compact prototypes and establishes prototype-feature similarity. This imposes implicit and hierarchical constraints that bridge task and representation gaps. Extensive experiments across four camouflaged and four underwater object benchmarks, under three degradation settings, demonstrate that our method consistently outperforms state-of-the-art approaches, highlighting its robustness and generalization under distribution shifts.
\end{abstract}


\section{Introduction}
Existing segmentation and detection frameworks have made notable progress in perceiving salient entities, yet they struggle with concealed targets. When objects with highly similar appearances blend into surrounding contexts without clear boundaries, they may critically mislead downstream analysis, \textit{e.g.,} medical image diagnosis \cite{fan2020pranet}. Camouflaged object detection (COD) focuses on identifying and localizing such hidden targets, providing critical visual cues for robust scene understanding and reliable decision-making.

\begin{figure*}[tb]
\includegraphics[width=1\textwidth]{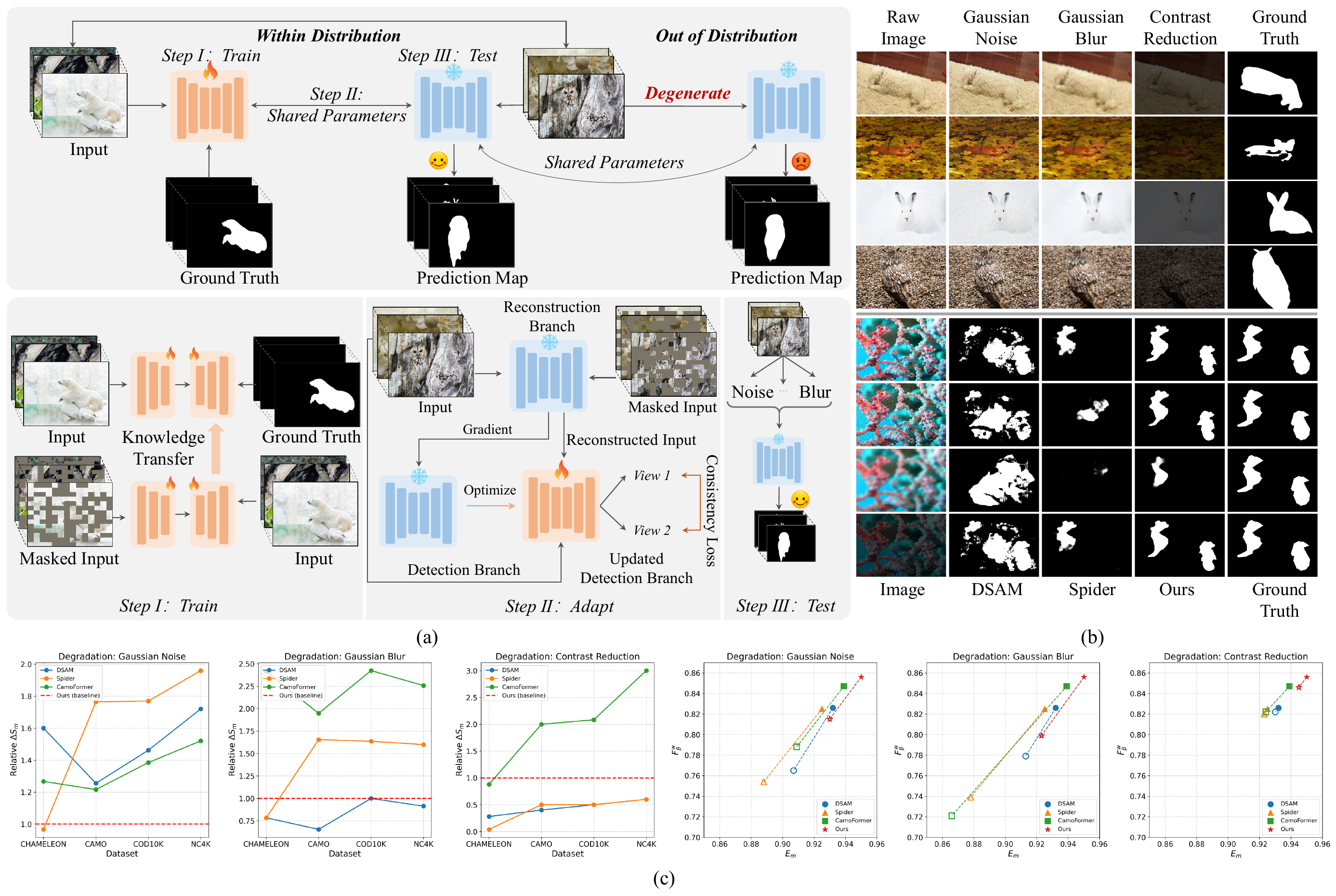}
\caption{(a) Traditional methods follow the \textit{train-fix-test} paradigm (top), while our method adopts the \textit{train-adapt-test} paradigm to dynamically perceive the scene (bottom). (b) Three testing sample degradation strategies (top) and qualitative comparisons (bottom). (c) Left: Performance drop $\Delta S_m$ of comparison methods, relative to our method. Right: Performance variation ($F^{w}_{\beta}$ and $E_m$ on the NC4K) before (solid markers) and after degradation (hollow markers). Best
view by zooming in.}
\label{fig:teaser}
\end{figure*}

COD presents fundamental challenges due to the strong visual entanglement between targets and backgrounds, \textit{e.g,} texture mimicry and chromatic assimilation. In Figure \ref{fig:teaser} (a), traditional approaches mainly adhere to a \textit{train-then-freeze} paradigm, where models are trained on offline datasets and deployed with fixed parameters. However, this paradigm exhibits significant limitations in open-world or degenerate scenarios (Figure \ref{fig:teaser} (b)): 1) \textbf{Domain Rigidity:} Pre-trained models lack the flexibility to adapt to diverse test-time scenes (\textit{e.g.,} variations in illumination), resulting in feature misalignment and poor pixel-level discrimination. 2) \textbf{Annotation Dependency:} Heavy reliance on limited labeled training data restricts generalization, particularly in long-tail or evolving camouflage scenarios, where exhaustive annotation is infeasible. Test-time adaptation (TTA) \cite{sun2020test, niu2022efficient} offers a \textit{train-then-adapt} paradigm that actively leverages intrinsic signals during inference. Rather than treating testing as a passive process, TTA enables models to dynamically recalibrate feature sensitivity, amplifying subtle yet crucial cues for camouflaged object perception. This shift allows models to self-optimize in the presence of distribution shifts. Motivated by this, we aim to formulate a task-specific TTA strategy for camouflage perception. However, seamlessly adapting existing TTA techniques to the COD task is non-trivial. Since camouflaged targets intentionally obscure boundaries and mimic background textures, directly applying standard self-supervised adaptation objectives often struggles to extract reliable discriminative representations. This raises three crucial questions regarding the conflicts between adaptation and camouflage characteristics:

\textit{How can we extract reliable self-supervised signals from heavily entangled multi-scale features?} In COD, visual ambiguity arises from intrinsic multi-scale feature entanglement, where local texture similarities and global attribute alignment blur the distinction between targets and backgrounds. Existing TTA methods often rely on single-modality reconstruction in the spatial domain, attempting to differentiate targets through pixel-level autoencoding. However, this approach suffers from critical limitations: it is highly sensitive to local perturbations (\textit{e.g.,} lighting variations), making it prone to overfitting on noise. Additionally, it decouples low-level details from high-level semantics, lacking robustness against occlusions and deformations that break spatial continuity and hinder global semantic reasoning. In camouflage scenarios, targets and backgrounds typically share predominant low-frequency features, with subtle differences hidden within high-frequency structural residuals. To address this, we propose a hierarchical feature disentanglement strategy through multi-spectral joint reconstruction. This method decomposes visual features in the spectrum space, suppressing high-frequency noise while modeling coarse target-background discrepancies. At the same time, it isolates and sharpens local details by enhancing edges and subtle textures, compensating for spatial ambiguity. Leveraging complementary constraints enables the model to disentangle complex visual patterns across different scales.

\textit{How can we bridge the optimization gap between self-supervised signals and main detection objectives to prevent learning drift?} During dynamic adaptation, self-supervised objectives inherently focus on overall image restoration. When relying solely on reconstruction loss to adjust model parameters during the test phase, the auxiliary reconstruction branch may overfit to local restoration. Meanwhile, the main detection branch, lacking direct task-specific supervision, gradually drifts away from the discriminative representations. This leads to a severe decoupling between the extracted self-supervised signals and the ultimate detection objectives. To prevent this, we propose an explicit cross-branch knowledge-sharing mechanism. Rather than treating reconstruction and detection as independent processes, this approach transforms structural-semantic correlations learned during joint training into dynamic constraints. Concretely, structural anomalies captured by the reconstruction branch, e.g., edge discontinuity or illumination inconsistency, are leveraged through an affinity matrix to dynamically correct the local feature responses of the detection branch. By explicitly mapping these structural anomalies to refine semantic features, we ensure that the adaptation process remains firmly anchored to the primary goal of distinguishing the concealed target.

\textit{How can we enforce semantic invariance to prevent the adapted model from collapsing into local details?} The discriminative core of camouflaged objects often lies in global semantic relationships rather than isolated local feature differences. Unconstrained pixel-level adaptation may introduce local noise or semantic deviations, especially when dealing with severe occlusions or heavy boundary blurring. Updating parameters based solely on local variations may lead the model to mistakenly assimilate target parts into the background. Guiding the model to ignore superficial perturbations and focus on high-level semantic invariance is therefore crucial. To achieve this, we construct a robust semantic representation space that shifts the optimization objective from pixel-level detail recovery to semantic consistency. By aggregating features from regions into compact prototype vectors and establishing similarity alignments, this approach encourages the model to suppress local disturbances and focus on consistent semantic cues shared across samples. Dynamically balancing local reconstruction with global semantic coherence prevents the model from collapsing into noise, thereby significantly enhancing generalization to unseen camouflage patterns.

Technically, we introduce the HRR to enforce pixel-level and multi-frequency reconstruction while imposing domain consistency constraints. To further enhance knowledge transfer, we propose the TAG, which constructs channel-wise affinity maps to propagate local knowledge from the reconstruction branch to the main detection branch. To prevent the model from collapsing into trivial details, we formulate the PCC, which feeds the reconstructed image into the detection model to generate predictions. We employ entropy estimation and edge-guided mechanisms to produce confidence maps, which emphasize regions of high uncertainty and object boundaries. We then perform variational fusion of prototypes from the original and reconstructed images, followed by metric consistency computation to ensure robust and reliable predictions. In Figure \ref{fig:teaser} (c), our method yields promising results on the original benchmark and under diverse conditions.

In summary, our main contributions are as follows:
\begin{itemize}
\setlength{\itemsep}{1pt}
\setlength{\parsep}{1pt}
\setlength{\parskip}{1pt}
\item We revisit existing frameworks for the COD task and propose a sample-specific TTA strategy to handle data distribution shifts. This strategy requires no additional data and can be seamlessly integrated into other methods.
\item We introduce three customized components: the HRR, which enables automatic adaptation to diverse scenes and imaging conditions; the TAG, which serves as a constraint to guide attention toward structurally inconsistent regions and refine features; and the PCC, which facilitates selective learning of decoupled and compact representations.
\item Extensive experiments on eight benchmarks and three distribution shift settings validate the superiority of the proposed method and the effect of components.
\end{itemize}

\section{RELATED WORK}
\subsection{Saliency Perception}
In contrast to camouflage perception, saliency perception aims to identify visually distinctive content. Based on feature selection strategies, existing methods can be broadly categorized into handcrafted and learning-based approaches. The former leverages expert prior knowledge, such as gradient or geometric cues, while the latter adopts data-driven paradigms to generate high-dimensional latent representations with greater robustness. Learning-based methods evolved beyond conventional 2D natural scenes to diverse downstream scenarios, including underwater, mirror, and panoramic environments \cite{zha2024dual,zha2025heterogeneous,zha2024weakly,zhao2023distortion,zha_aaai_2026}, and have progressed from single-modality frameworks to multi-modal collaboration (\textit{e.g.,} depth and thermal data). Furthermore, some works focus on modeling relative saliency differences among targets, \textit{i.e.,} saliency ranking \cite{liu2025language}.
Since fully supervised learning generally requires large-scale, multi-scene datasets to obtain generalizable representations, recent efforts explored data-efficient alternatives such as self-supervised learning \cite{guan2025contrastive}.
In this work, we propose to discover discriminative regions during inference by leveraging the image itself as a supervisory signal.

\subsection{Camouflage Perception} The development of COD has progressed rapidly. Based on supervised learning signals, we categorize it into four types: 1) Fully supervised, which utilizes edge \cite{sun2022boundary}, texture \cite{zhu2021inferring}, frequency \cite{sun2024frequency}, depth \cite{wu-popNet}, uncertainty \cite{zhang2023predictive}, or text \cite{zhang2024unlocking} guidance, or employs a coarse-to-fine progressive approach to mine potential clues \cite{du2025upgen}; 2) Weakly supervised \cite{chen2024just,zha2026think}, which leverages manually annotated scribbles or points as ground truth and iteratively optimizes the labels; 3) Semi-supervised \cite{lai2024camoteacher}, which explores the relationships and consistencies between labeled and unlabeled data; 4) Zero/Few-shot learning \cite{li2023zero,wang2024few}, which aims to transfer knowledge to unseen scenarios using only a limited number of samples. Additionally, some works \cite{zhao2024focusdiffuser,chen2024sam,luo2024vscode,zhang2025comprompter} leverage features and semantic priors from pre-trained or large language/vision/multi-modal foundation models, introducing efficient fine-tuning mechanisms that achieve competitive performance with minimal learnable parameters. Furthermore, some studies investigate at the instance level \cite{he2024text}, video level \cite{hui2024endow,zhang2025explicit}, and across different scenarios \cite{wang2024depth}. Our work is most closely related to \cite{hao2025simple}, but it overlooks scene adaptation during the testing phase by constructing a multi-task framework for pixel reconstruction and detection that is susceptible to noise interference. Pang \textit{et al.} \cite{pang2024open} proposed an open-world setting based on CLIP \cite{radford2021learning}, utilizing text prompts. In contrast, our HCL does not rely on additional data or foundation models to facilitate the remapping of the model to address distribution shifts.

\subsection{Test-time Adaptation} TTA adapts model parameters during the testing phase to dynamically align with new data distributions, facilitating the transition from \textit{i.i.d.} modeling to out-of-distribution generalization. Sun \textit{et al.} \cite{sun2020test} proposed adapting models via online self-supervised learning, enabling real-time updates. Wang \textit{et al.} \cite{wang2020tent} further improved efficiency by dynamically adjusting parameters via entropy minimization. To tackle error accumulation, Niu \textit{et al.} \cite{niu2022efficient} introduced anti-forgetting mechanisms with sample selection and regularization. Jang \textit{et al.} \cite{jang2022test} enhanced pseudo-label reliability by leveraging prototype matching. For lifelong adaptation, Brahma \textit{et al.} \cite{brahma2023probabilistic} designed a probabilistic framework to balance new knowledge integration with prior preservation. Zhao \textit{et al.} \cite{zhao2023delta} tackled complex distribution shifts by combining statistical correction and sample reweighting. Ma \textit{et al.} \cite{ma2024improved} boosted robustness through graph-structured label refinement and model averaging. Recently, some works introduced TTA into specific tasks, \textit{e.g.,} video segmentation \cite{liu2024depth}. We introduce TTA into the COD and propose the HCL framework, which incorporates tailored components to mitigate the train-test gap, particularly under significant distribution shifts. Unlike \cite{hu2024relax}, which relies on multiple foundation models and adapts only at the test stage through prompting, the HCL framework integrates seamlessly into the entire pipeline without requiring manual design (\textit{e.g.,} prompts) or prior external knowledge.

\section{PROPOSED METHOD}

\subsection{Overall Architecture}
\begin{figure*}[tb]
  \centering
  \includegraphics[width=1\textwidth]{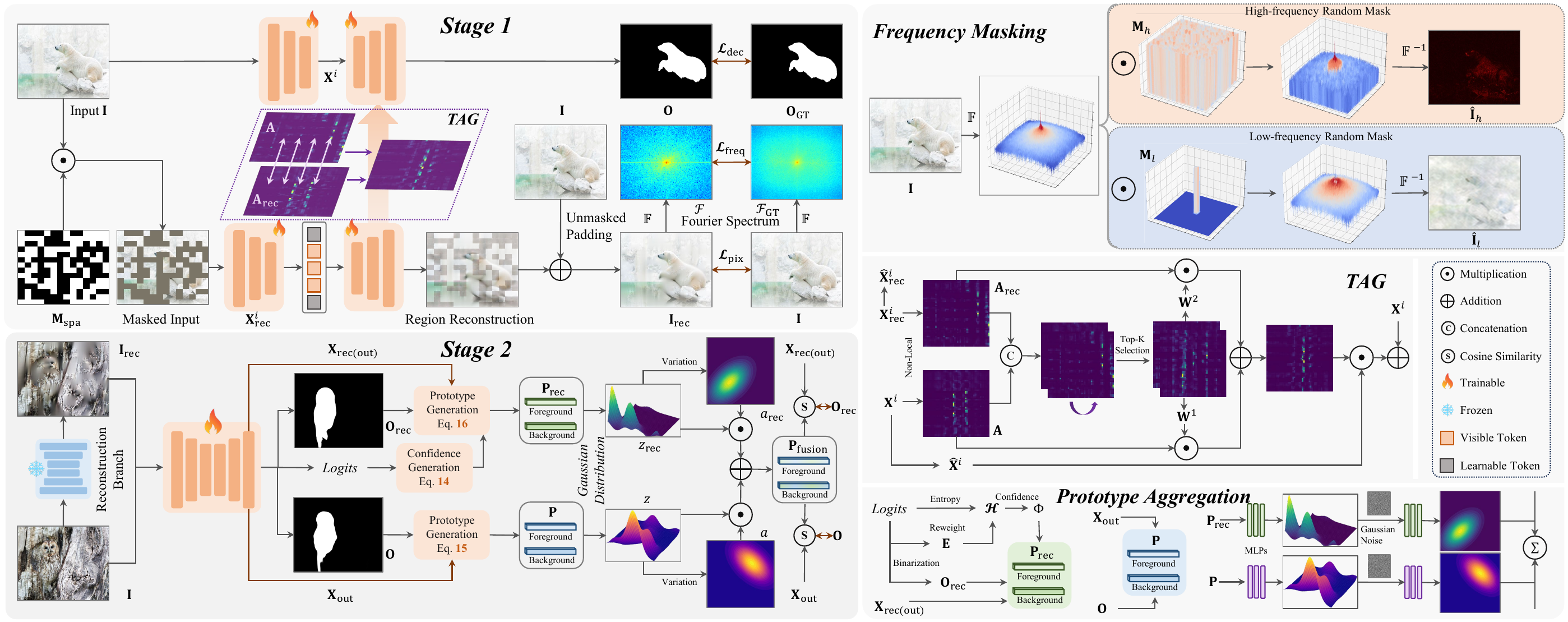}
  \caption{Stage 1: The input image undergoes masked patch partitioning in the spatial domain, while threshold-based masking is applied to obtain high- and low-frequency components. Unmasked patches are encoded to extract visible tokens, and along with learnable tokens, are decoded in spatial and frequency branches for pixel- and spectrum-level reconstruction. The TAG enables cross-branch knowledge transfer, serving as an explicit constraint. Stage 2: The spatially reconstructed image is passed through the detection network to generate predictions. Uncertainty estimation and edge information are leveraged to construct the confidence map that refines pixel-wise weighting. Finally, prototype integration enforces cross-view consistency, forming a closed-loop from reconstruction to prediction.}
  \label{Framework}
\end{figure*}

In Figure \ref{Framework} and Algorithm \ref{alg:core}, given an input $\mathbf{I} \in \mathbb{R}^{3 \times H \times W}$, we generate a random mask $\mathbf{M}_{\rm spa} \in \{0,1\}^{H \times W}$ and extract multi-scale features $\mathbf{X}^{i} \in \mathbb{R}^{C^i \times H^i \times W^i}$ using the encoder ($C$: Channel; $H$: Height; $W$: Width). We apply Fourier transform $\mathbb{F}$ to project the input into the frequency domain, where mask $\mathbf{M}_{l}$ (and $\mathbf{M}_{h}$) and inverse transformation $\mathbb{F}^{-1}$ are used to obtain the masked low-frequency component $\hat{\mathbf{I}}_{l}$ and the masked high-frequency component $\hat{\mathbf{I}}_{h}$. The subsequent operations mirror those in the spatial domain. We alternately use $\mathcal{L}_{\rm pix}$ and $\mathcal{L}_{\rm freq}$ as reconstruction and constraint losses. The masked feature $\mathbf{X}_{\rm rec}^{i}$ transfers heterogeneous information to the unmasked feature $\mathbf{X}^{i}$ via the TAG as a bridge. The reconstructed image $\mathbf{I}_{\rm rec}$ is then fed as an additional input to generate the prediction map $\mathbf{O}_{\rm rec}$ and confidence map $\Phi$, which are further used to derive the prototype $\mathbf{P}_{\rm rec}$. We integrate $\mathbf{P}_{\rm rec}$ with the prototype $\mathbf{P}$ from the original input, sequentially using the fused prototype $\mathbf{P}_{\rm fusion}$ as an anchor to establish consistency across predictions and features.

\subsection{Preliminary}
According to Parseval's Theorem, a signal $x$ retains the same total energy in both the spatial and frequency domains, though with different emphasis. Based on the 2D Discrete Fourier Transform (DFT) $\mathbb{F}$, we transform $x \in \mathbb{R}^{H \times W}$ into the Fourier spectrum,
\begin{equation}
\setlength\abovedisplayskip{3pt}
\setlength\belowdisplayskip{3pt}
\mathcal{F}(x)(u, v) = \sum_{h=0}^{H-1} \sum_{w=0}^{W-1} x(h, w) e^{-j2\pi \left( \frac{h}{H} u + \frac{w}{W} v \right)}
\end{equation}
where $u$ and $v$ represent the horizontal and vertical indices in the spectrum, respectively. To transform back to the spatial domain, we apply the inverse Fourier transform $\mathbb{F}^{-1}$,
\begin{equation}
\setlength\abovedisplayskip{3pt}
\setlength\belowdisplayskip{3pt}
x(h, w) = \frac{1}{HW} \sum_{u=0}^{H-1} \sum_{v=0}^{W-1} \mathcal{F}(u, v) e^{j2\pi \left( \frac{uh}{H} + \frac{vw}{W} \right)}
\end{equation}

In practice, we use the Fast Fourier Transform (FFT) version to improve computational efficiency.

\subsection{Hierarchical Representation Reconstruction}
Reconstruction is not a fundamental component of camouflage perception in terms of task definition. Instead, we employ it to obtain robust and effective self-supervised learning signals. We decompose the auxiliary reconstruction task into: spatial, high-frequency, and low-frequency, each accompanied by corresponding transformation constraints. 

\noindent\textbf{Spatial Reconstruction.} To supervise the reconstruction process, we adopt the Mean Square Error (MSE) loss $\mathcal{L}_{\mathrm{pix}}$ on the output $\mathbf{I}_{\rm rec}$ from the pixel decoder, encouraging accurate recovery of the spatial regions masked by $\mathbf{M}_{\rm spa}$,
\begin{equation}
\setlength\abovedisplayskip{3pt}
\setlength\belowdisplayskip{3pt}
\mathcal{L}_{\mathrm{pix}} = \frac{1}{HWC}\sum_{i=0}^{H-1} \sum_{j=0}^{W-1} \sum_{c=0}^{C-1} \left( \mathbf{I}_{\rm rec}^{i,j,c} - \mathbf{I}^{i,j,c} \right)^2
\end{equation}

\noindent\textbf{Frequency Reconstruction.} To avoid inherent reconstruction defects in the pixel domain, we employ the Fourier transform to convert the raw image into the frequency domain, partitioning the spectrum into high- and low- frequency part based on the predefined threshold (the distance of spectral points from the center). For low-frequency reconstruction, we use a central mask $\mathbf{M}_{l} \in \{0,1\}^{H_l \times H_w}$ to randomly suppress low-frequency energy while retaining high-frequency, 
\begin{equation}
\setlength\abovedisplayskip{3pt}
\setlength\belowdisplayskip{3pt}
% \mathbf{M}(h, w) \ \texttt{or} \ 
\mathbf{M}_{l}(u, v) = 
\begin{cases}
1, & \texttt{unmasked} \\
0, & \texttt{masked}
\end{cases}
\end{equation}

Therefore, we can obtain the degraded spectrum $\hat{\mathcal{F}}$, 
\begin{equation}
\setlength\abovedisplayskip{3pt}
\setlength\belowdisplayskip{3pt}
\hat{\mathcal{F}} = \sum_{c=0}^{C-1} \mathcal{F}_{\rm low}^{c} \odot \mathbf{M}_{l}^{c} + \mathcal{F}_{\rm high}^{c}
\end{equation}
where $\odot$ denotes the Hadamard product. We further utilize the inverse Fourier transform to generate the degraded image $\hat{\mathbf{I}}_{l}$. For high-frequency reconstruction, we perform similar operations and define the random mask $\mathbf{M}_{h}$, obtaining $\hat{\mathbf{I}}_{h}$. After generating features using the encoder, we reconstruct the high-frequency and low-frequency images from tokens, \textit{i.e.,} $\mathbf{\mathbf{I}_{\rm rec (h)}}$ and $\mathbf{\mathbf{I}_{\rm rec (l)}}$.

In spatial visual recognition, different regions contribute unequally to decision-making. Likewise, in frequency decoding, hard samples demand greater learning attention. To address this, we introduce focal frequency loss \cite{jiang2021focal} to capture high-value clues and emphasize challenging frequency components,
\begin{equation}
\setlength\abovedisplayskip{3pt}
\setlength\belowdisplayskip{3pt}
\mathcal{L}_{\mathrm{freq}} = \frac{1}{HW} \sum_{u=0}^{H-1} \sum_{v=0}^{W-1} \omega(u, v) \odot \gamma(\mathcal{F}(u, v), \mathcal{F}_{\rm GT}(u, v))^2
\end{equation}
where $\mathcal{F}(u, v)$ and $\mathcal{F}_{\rm GT}(u, v)$ represent the components of the reconstructed spectrum $\mathbf{L}_{\rm rec}$ (or $\mathbf{H}_{\rm rec}$) and the raw spectrum $\mathbf{L}$ (or $\mathbf{H}$) at position $(i,j)$, respectively. $w$ and $\gamma$ denote the weight parameter and distance metric, which can be formulated as,
\begin{equation}
\setlength\abovedisplayskip{3pt}
\setlength\belowdisplayskip{3pt}
\begin{split}
\omega(u, v) &= \gamma(\mathcal{F}(u, v), \mathcal{F}_{\rm GT}(u, v))^{\beta} \\
\gamma(\mathcal{F}, \mathcal{F}_{\rm GT}) &= \sqrt{(\mathcal{R} - \mathcal{R}_{\rm GT})^2 + (\mathcal{I} - \mathcal{I}_{\rm GT})^2}
\end{split}
\end{equation}
where $\mathcal{R}$ and $\mathcal{I}$ denote the real and imaginary parts of the spectrum, respectively. $\beta$ is a scaling factor, which is set to 1 by default.

\noindent\textbf{Transformation Consistency.} In theory, when $\mathcal{L}_{\rm pix} \to 0$ or $\mathcal{L}_{\rm freq} \to 0$, the model can accurately predict the masked regions, resulting in the reconstructed image that matches the raw image. However, $\mathcal{L}_{\rm pix}$ directly enforces pixel-level matching, ensuring the recovery of local details but is insensitive to global structures, \textit{e.g.,} edge continuity. Conversely, $\mathcal{L}_{\rm freq}$ enhances the perception of global structures but is less sensitive to local pixel shifts. In other words, a sole focus on pixel matching may lead to blurriness, while an emphasis on spectral matching may introduce artifacts. Therefore, for each reconstruction branch, we apply the corresponding domain loss as a regularization term to achieve transformation consistency. We can formulate the total loss $\mathcal{L}_{\rm HRR}$ for the HRR,
\begin{align}
\setlength\abovedisplayskip{3pt}
\setlength\belowdisplayskip{3pt}
\mathcal{L}_{\rm HRR} &= \mathcal{L}_{\rm{pix} (rec)}(\mathbf{I}_{\rm rec}, \mathbf{I}) + \mathcal{L}_{\rm{freq (con)}}(\mathcal{F},\mathcal{F}_{\rm GT}) \nonumber \ + \\
&  \sum_{k \in \{\mathbf{L}, \mathbf{H}\}} \lambda_k \left( \mathcal{L}_{\rm{freq (rec)}}(k_{\rm rec}, k) + \mathcal{L}_{\rm{pix} (con)}(\mathbb{F}^{-1}(k),\mathbf{I}) \right)
\end{align}
where $\mathcal{L}_{\sim \rm (rec)}$ and $\mathcal{L}_{\sim \rm (con)}$ represent the reconstruction and consistency (constraint) losses, respectively. 
$\lambda$ is the balancing hyperparameter. We empirically set $\lambda_{\mathbf{L}}$ and $\lambda_{\mathbf{H}}$ to 0.4 and 0.6.

\subsection{Task Affinity Guidance}
Despite employing reconstruction as auxiliary supervision and implicitly updating the detection model parameters via gradient optimization, our approach may require more training iterations and could lead to suboptimal performance. To address this limitation, we explicitly establish feature associations. Unlike vanilla cross-attention mechanisms with high computational complexity ($\mathcal{O}(H^2W^2)$), we construct a more efficient affinity matrix to reduce computational overhead and suppress noise. Specifically, we adopt channel-wise non-local associations with reduced complexity ($\mathcal{O}(C^2)$) to capture long-range dependencies and enhance informative representations, ultimately generating the affinity map $\mathbf{A}$ and the optimized feature $\hat{\mathbf{X}}$,
\begin{equation}
\setlength\abovedisplayskip{3pt}
\setlength\belowdisplayskip{3pt}
[\hat{\mathbf{X}}, \mathbf{A}] = \texttt{NonLocal}(\mathbf{X})
\end{equation}

For $\mathbf{X}_{\rm rec}$, the operation follows the same procedure, producing $\hat{\mathbf{X}}_{\rm rec}$ and $\mathbf{A}_{\rm rec}$. Due to representation differences, we fuse $\mathbf{A}$ and $\mathbf{A}_{\rm rec}$ to obtain the weight map $\mathbf{W}$,
\begin{equation}
\setlength\abovedisplayskip{3pt}
\setlength\belowdisplayskip{3pt}
\small
[\mathbf{W}^1, \mathbf{W}^2] = \texttt{Top-K}(\texttt{Self-Attention}(\texttt{Concat}(\mathbf{A}, \mathbf{A}_{\rm rec})))
\end{equation}
where $\texttt{Concat}$ denotes channel concatenation, while $\texttt{Top-K}$ selects the top-k weights to filter out low-response elements. In Figure \ref{fig:Hyperparameter settings}, we select the top 70\% (or 80\%) of elements to achieve a high signal-to-noise ratio. Thus, we can obtain the updated map that incorporates the reconstruction knowledge,
\begin{equation}
\setlength\abovedisplayskip{3pt}
\setlength\belowdisplayskip{3pt}
\mathbf{A} := \mathbf{W}^1 \odot \mathbf{A} + \mathbf{W}^2 \odot \mathbf{A}_{\rm rec}
\end{equation}

Correspondingly, the updated feature is, 
\begin{equation}
\setlength\abovedisplayskip{3pt}
\setlength\belowdisplayskip{3pt}
\hat{\mathbf{X}} :=  \mathbf{X} + \mathbf{A} \odot \hat{\mathbf{X}}
\end{equation}

\subsection{Prototype
Consistency Calibration}
% \textbf{Reliability Map Generation.}
% \cite{rubinstein2016simulation}
\noindent\textbf{Confidence Map Generation.} In practice, the representations of the reconstruction and the original perspective cannot be fully identical. Inevitably, noise disturbances and differentiated modeling can lead to unreliability in certain regions of the predicted map $\mathbf{O}_{\rm rec}$. To quantify the degree of reliability, we introduce uncertainty estimation. Unlike Monte Carlo strategy that requires multiple samples, we measure based on entropy $\mathcal{H}$, which demands only a single forward pass. For camouflaged targets, the boundary information is more critical than the main body. Therefore, we generate the edge map $\mathbf{E}$ and utilize it as a spatial weighting factor. For any position $(i,j)$ of output-layer feature $\mathbf{X}_{\rm rec (out)} \in \mathbb{R}^{1 \times H \times W}$, we apply sigmoid to obtain the  probability $\mathbf{p}$ for class $m$,
\begin{equation}
\setlength\abovedisplayskip{3pt}
\setlength\belowdisplayskip{3pt}
\begin{split}
\mathcal{H}(\mathbf{p}^{i,j}) &= - \sum_{m=0}^{M-1} w(i,j) \mathbf{p}^{i,j}(m) \log \mathbf{p}^{i,j}(m) \\
w(i, j) &= 1 + \alpha \cdot \mathbf{E}(i, j)
\end{split}
\end{equation}
where $\alpha$ is a scaling factor to balance pixel-wise weights between edge and non-edge regions, and $\mathbf{E}(i,j) \in \{0,1\}$ denotes the edge indicator function: $\mathbf{E}(i,j)=1$ if pixel $(i,j)$ lies on generated edge, and 0 otherwise (for COD task, $M=2$). In information theory, entropy serves as a quantitative measure of uncertainty, where higher entropy values correspond to increased information disorder and probabilistic ambiguity in the prediction confidence. Based on this, we further define the confidence map $\Phi$,
\begin{equation}
\setlength\abovedisplayskip{3pt}
\setlength\belowdisplayskip{3pt}
\Phi^{i,j} = \frac{1}{H \times W} \left( 1 - \frac{\mathcal{H}(\mathbf{p}^{i,j})}{\sum_{i=0}^{H-1}\sum_{j=0}^{W-1} \mathcal{H}(\mathbf{p}^{i,j})} \right)
\end{equation}

\noindent\textbf{Prototype Generation.} The prototype characterizes the overall representation of a region. We generate the raw prototype via masked average pooling (MAP),
\begin{equation}
\setlength\abovedisplayskip{3pt}
\setlength\belowdisplayskip{3pt}
\mathbf{P}^m = \frac{\sum_{i=0}^{H-1}\sum_{j=0}^{W-1} \mathbf{X}_{\rm out}^{i,j} \mathds{1}[\mathbf{O}^{i,j} = m]}{\sum_{i=0}^{H-1}\sum_{j=0}^{W-1}  \mathds{1}[\mathbf{O}^{i,j} = m]}
\end{equation}
where $\mathds{1}$ is an indicator function. MAP treats features from all regions equally. However, the reconstructed features may contain errors (\textit{e.g.,} blurriness and noise), particularly in incomplete or complex areas. Directly using MAP may lead to the smoothing of errors or even obscure the correct representations. We utilize $\Phi$ to automatically filter reliable samples, achieving self-calibration,
\begin{equation}
\setlength\abovedisplayskip{3pt}
\setlength\belowdisplayskip{3pt}
\mathbf{P}^m_{\rm rec} = \frac{\sum_{i=0}^{H-1}\sum_{j=0}^{W-1} \mathbf{X}^{i,j}_{\rm rec (out)} \Phi^{i,j} \mathds{1}[\mathbf{O}_{\rm rec}^{i,j} = m]}{\sum_{i=0}^{H-1}\sum_{j=0}^{W-1} \mathds{1}[\mathbf{O}_{\rm rec}^{i,j} = m]}
\end{equation}

\noindent\textbf{Variational Prototype Fusion.} While naive fusion strategies like element-wise summation or channel-wise concatenation between prototypes $\mathbf{P}^m$ and $\mathbf{P}^m_{\rm rec}$ appear straightforward, both suffer from critical limitations: 1) Direct summation induces linear amplification of noise artifacts: if $\mathbf{P}^m$ contains misclassified background pixels due to blurred object boundaries and $\mathbf{P}^m_{\rm rec}$ exhibits restoration distortions (\textit{e.g.,} occlusion),  error patterns become compounded via accumulation; 2) Concatenation risks feature space misalignment, as $\mathbf{P}^m$  primarily encodes fine-grained representations while $\mathbf{P}^m_{\rm rec}$ captures structural knowledge, creating distribution shifts that hinder semantic consistency when forcibly combined. We utilize variational fusion to alleviate through probabilistic modeling.

To further explore the intrinsic uncertainty, we project the samples into the latent Gaussian distribution $\mathcal{N}$ to establish semantic associations. For $\mathbf{P}^m$, we have, 
\begin{equation}
\setlength\abovedisplayskip{3pt}
\setlength\belowdisplayskip{3pt}
\begin{split}
&p(z|\mathbf{P}^m) = \mathcal{N}(z; \mu, (\sigma)^2 \mathcal{I}) \\
&\mu = \mathcal{M}_{\mu}(\mathbf{P}^m), \ \sigma = \mathcal{M}_{\sigma}(\mathbf{P}^m)
\end{split}
\end{equation}
where $z$, $\mu$, $\sigma$, and $\mathcal{I}$ represent the reconstruction vector, mean, variance, and identity matrix, respectively. A larger $\sigma$ indicates greater uncertainty. $\mathcal{M}$ denotes the multilayer perceptron (MLPs). Similarly, for $\mathbf{P}^m_{\rm rec}$, we can obtain $z_{\rm rec}$.

Sampling latent variables from a distribution to generate attention weights results in non-differentiable gradients. To resolve this, we employ the reparameterization trick, which decouples the stochasticity by introducing external noise, allowing gradient flow through a deterministic computational path. Specifically, we introduce standard Gaussian noise $\eta$, 
\begin{equation}
\setlength\abovedisplayskip{3pt}
\setlength\belowdisplayskip{3pt}
z = \mu + \eta \cdot \sigma, \quad \eta \in \mathcal{N}(0, 1)
\end{equation}

A similar operation is performed for $z_{\rm rec}$. Traditional attention mechanisms (like Softmax) generate fixed weights $a^{i}$ through point estimates, which are essentially deterministic mappings and cannot reflect the model's confidence in the weight values, 
\begin{equation}
\setlength\abovedisplayskip{3pt}
\setlength\belowdisplayskip{3pt}
\begin{split}
\hat{a}^i &= \mathbf{W}^{i} z^i + \mathbf{b}^i, \quad i \in \{o, r\} \\
a^i &= \frac{\exp(\hat{a}^i)}{\sum_{j \in \{o, r\}} \exp(\hat{a}^{j})}
\end{split}
\label{point}
\end{equation}
where $o$ and $r$ denote the original and reconstructed spatial images, respectively. $\mathbf{W}$ and $\mathbf{b}$ are learnable parameters used for projection. Instead of using deterministic weights, we leverage dual uncertainty modeling by treating the weights as probability distributions. \textit{1) Feature-level uncertainty:} The variances $\sigma$ and $\sigma_{\rm rec}$ of the prototype features represent local confidence. \textit{2) Weight-level uncertainty:} The variance $\sigma_a$ of the attention weights captures global confidence.

For each latent variable $z^i$, we employ shared MLPs to estimate the mean $\mu_{a}$ and variance $\sigma_{a}$. Accordingly, we reformulate Eq.~\ref{point} as follows,
\begin{equation}
\setlength\abovedisplayskip{3pt}
\setlength\belowdisplayskip{3pt}
\begin{split}
\hat{a}^i \sim q(\hat{a}^i \mid z^i) &= \mathcal{N}\left(\mu_a^i, (\sigma_a^i)^2 \mathbf{I}\right), \quad i \in \{o, r\} \\
a^i &= \frac{\exp \left( -\gamma (\sigma_a^i)^2 \right)}{\sum_{j \in \{o, r\}} \exp \left( -\gamma (\sigma_a^{j})^2 \right)}
\end{split}
\end{equation}
where $\gamma$ is a learnable temperature coefficient that controls the intensity of variance suppression on the weights. We can obtain the prototype $\mathbf{P}_{\rm fusion}^{m}$ after weighted fusion,
\begin{equation}
\setlength\abovedisplayskip{3pt}
\setlength\belowdisplayskip{3pt}
\mathbf{P}_{\rm fusion}^m = \sum_{i \in \{o, r\}} a^i z^i
\end{equation}

To prevent overfitting to noise, we enforce that the prototype distribution approaches the standard Gaussian prior through Kullback–Leibler (KL) divergence,
\begin{equation}
\setlength\abovedisplayskip{3pt}
\setlength\belowdisplayskip{3pt}
\mathcal{L}_{\rm{KL}} = \sum_{i \in \{o, r\}} \frac{1}{2} \left( (\mu^i)^2 + (\sigma^i)^2 - \log((\sigma^i)^2) - 1 \right)
\end{equation}

Based on variational fusion, we reduce the weights in high-variance (low confidence) regions to suppress noise propagation. We dynamically adjust the fusion strategy based on input scenarios to achieve adaptive feature complementarity.

\noindent\textbf{Metric Consistency.} We use the similarity matrix $\mathbf{S}$ between feature and prototype to estimate the class probability for each pixel. Essentially, this strategy is an implicit form of contrastive learning,
\begin{equation}
\setlength\abovedisplayskip{3pt}
\setlength\belowdisplayskip{3pt}
\mathbf{S}^{i,j} = \texttt{CosSim}(\mathbf{X}^{i,j}, \mathbf{P}^m) = \frac{\mathbf{X}^{i,j} \cdot \mathbf{P}^m}{\|\mathbf{X}^{i,j}\|_2 \cdot \|\mathbf{P}^m\|_2 + \epsilon}
\end{equation}
where $\texttt{CosSim}(\cdot)$ and $\epsilon$ represent the cosine similarity and the minimum value, respectively. Similarly, we can obtain $\mathbf{S}_{\rm rec}$ for $\mathbf{P}_{\rm rec}$. When $\mathbf{P}_{\rm fusion}$ is sufficiently accurate and robust, it follows that $\mathbf{S} \to \mathbf{O}$ (or $\mathbf{S}_{\rm rec} \to \mathbf{O}_{\rm rec}$). Base on cross entropy loss $\mathcal{L}_{\rm CE}$, we have,
\begin{equation}
\setlength\abovedisplayskip{3pt}
\setlength\belowdisplayskip{3pt}
% \small
\mathcal{L}_{\mathrm{pro}} = \mathcal{L}_{\mathrm{CE}}(\mathbf{S}, \mathbf{O}), \ \mathcal{L}_{\mathrm{pro (rec)}} = \sum_{i=0}^{H-1}\sum_{j=0}^{W-1} \Phi^{i,j} \mathcal{L}_{\mathrm{CE}}(\mathbf{S}_{\rm rec}^{i,j}, \mathbf{O}_{\rm rec}^{i,j})
\end{equation}

For the main detection branch, following \cite{pang2022zoom}, we employ structure loss $\mathcal{L}_{\rm dec}$. Note that we apply supervision to the predictions at all stages of the decoder. We can derive the total loss $\mathcal{L}_{\rm total}$,
\begin{equation}
\setlength\abovedisplayskip{3pt}
\setlength\belowdisplayskip{3pt}
\mathcal{L}_{\rm totoal} = \sum_{i \in {\rm HRR, \ KL, \ pro, \ proe(rec),\ dec}} \lambda_{i}\mathcal{L}_{i}
\end{equation}
% where $\lambda$ is the balancing parameter.

In the TTA phase, we discard $\mathcal{L}_{\rm dec}$.

\begin{algorithm}
\caption{Our HCL Framework}
\label{alg:core}
\begin{algorithmic}[1]
% \State \textbf{Input:} Image $\mathbf{I}$; \textbf{Output:} Prediction Map
\State \textbf{Input:} Image $\mathbf{I}$
\State \textbf{Stage 1: Hierarchical Reconstruction}
\State $\mathbf{M}_{\rm spa} \leftarrow \mathsf{SpaMask}(\textbf{I})$, $[\mathbf{M}_{l}, \mathbf{M}_{h}] \leftarrow \mathsf{FreqMask}(\mathbb{F}(\mathbf{I}))$
\State $\mathbf{X}_{\rm rec}, \mathbf{X} \leftarrow \mathsf{Enc}(\mathbf{I} \odot \mathbf{M}_{\rm spa}, \mathbf{I})$ \Comment{Encoding Process}
\State $[\mathbf{X}_{\rm rec(l)},\mathbf{X}_{\rm rec(h)}],\,\mathbf{X} \leftarrow \mathsf{Enc}\big(\mathbb{F}^{-1}(\mathbb{F}(\mathbf{I}) \odot [\mathbf{M}_l,\mathbf{M}_h]),\,\mathbf{I}\big)$
\State $\mathbf{X} \leftarrow \mathsf{TAG}(\mathbf{X}_{\rm rec}, \mathbf{X})$, $\mathbf{X} \leftarrow \mathsf{TAG}([\mathbf{X}_{\rm rec(l)},\mathbf{X}_{\rm rec(h)}], \mathbf{X})$
\State $\mathbf{O}, \mathbf{I}_{\rm rec} \leftarrow \mathsf{Dec}(\mathbf{X}_{\rm rec}, \mathbf{X})$ \Comment{Decoding Process}
\State $\mathbf{O}, [\mathbf{I}_{\rm rec(l)},\mathbf{I}_{\rm rec(h)}] \leftarrow \mathsf{Dec}(\mathbf{X}_{\rm rec}, \mathbf{X})$

\State \textbf{Stage 2: Consistency Calibration}
\State $\mathbf{I}_{\rm rec} \leftarrow \mathsf{FrozenNet}(\mathbf{I})$ \Comment{Reconstruction} 
\State $[\mathbf{X}_{\rm rec(out)}, \mathbf{O}_{\rm rec},\Phi] \leftarrow \mathsf{TrainableNet}(\mathbf{I}_{\rm rec})$ \Comment{Detection}
\State $[\mathbf{X}, \mathbf{O}] \leftarrow \mathsf{TrainableNet}(\mathbf{I})$
\State $\mathbf{P}_{\rm rec}, \mathbf{P} \leftarrow \mathsf{Prototype([\mathbf{X}_{\rm rec(out)}, \mathbf{O}_{\rm rec},\Phi],[\mathbf{X}, \mathbf{O}])}$
\State $\mathbf{P}_{\rm fusion} \leftarrow \mathsf{VariationalFusion}(\mathbf{P}_{\rm rec}, \mathbf{P})$
\State $\mathbf{S}_{\rm rec}, \mathbf{S} \leftarrow \mathsf{CosSim}(\mathbf{P}_{\rm fusion}, [\mathbf{X}_{\rm rec(out)}, \mathbf{X}_{\rm out}])$

\If{NOT INFERENCE}
    \State $\mathcal{L}_{\rm total} \leftarrow \mathcal{L}_{\rm HRR} + \mathcal{L}_{\rm KL} + \mathcal{L}_{\rm pro} + \mathcal{L}_{\rm pro (rec)} + \mathcal{L}_{\rm dec}$
\Else
    \State $\mathcal{L}_{\rm total} \leftarrow \mathcal{L}_{\rm HRR} + \mathcal{L}_{\rm KL} + \mathcal{L}_{\rm pro} + \mathcal{L}_{\rm pro (rec)}$ 
\EndIf
\State \textbf{Output:} Prediction Map $\mathbf{O}$
\end{algorithmic}
\end{algorithm}

\begin{table*}[tb]
\caption{Quantitative comparison on normal benchmark datasets. Foundation models include, \textit{e.g.,} SAM. Extra data includes, \textit{e.g.,} depth map and more pixels. Best performance in \textbf{bold}, second in \underline{underline}. $\ddagger$ represents data is unavailable. $\dagger$ denotes using HCL. *: Higher/Multi-scale input resolution. $\uparrow$ indicates higher values are better, while $\downarrow$ indicates the opposite. `Base-R' and `Base-P' denote ResNet50 and PVT-v2 as backbone, respectively.}
\centering
\resizebox{\linewidth}{!}{
\setlength\tabcolsep{2.5pt}
\begin{tabular}{l|c|c|c|cccc|cccc|cccc|cccc}
\toprule
\multirow{2}{*}{Method} 
& \multicolumn{1}{c|}{\multirow{2}{*}{\begin{tabular}{@{}c@{}}Foundation \\ Model\end{tabular}}} 
& \multicolumn{1}{c|}{\multirow{2}{*}{\begin{tabular}{@{}c@{}}Extra \\ Data\end{tabular}}} 
& \multicolumn{1}{c|}{\multirow{2}{*}{TTA}} 
& \multicolumn{4}{c|}{CHAMELEON (76)}
& \multicolumn{4}{c|}{CAMO (250)}
& \multicolumn{4}{c|}{COD10K (2026)}
& \multicolumn{4}{c}{NC4K (4121)} \\
\cmidrule(lr){5-8} \cmidrule(lr){9-12} \cmidrule(lr){13-16} \cmidrule(lr){17-20}
& & & & $S_m \uparrow$ & $E_m \uparrow$ & $F^{w}_{\beta} \uparrow$ & MAE $\downarrow$
& $S_m \uparrow$ & $E_m \uparrow$ & $F^{w}_{\beta} \uparrow$ & MAE $\downarrow$
& $S_m \uparrow$ & $E_m \uparrow$ & $F^{w}_{\beta} \uparrow$ & MAE $\downarrow$
& $S_m \uparrow$ & $E_m \uparrow$ & $F^{w}_{\beta} \uparrow$ & MAE $\downarrow$ \\
\midrule
PraNet \cite{fan2020pranet} \textsubscript{MICCAI'20}
& \ding{55} & \ding{55} & \ding{55} 
& 0.860 & 0.898 & 0.763 & 0.044
& 0.769 & 0.833 & 0.663 & 0.094
& 0.789 & 0.839 & 0.629 & 0.045
& 0.822 & 0.876 & 0.724 & 0.059 \\
SINet \cite{fan2020camouflaged} \textsubscript{CVPR'20}
& \ding{55} & \ding{55} & \ding{55} 
& 0.872 & 0.946 & 0.806 & 0.034
& 0.751 & 0.771 & 0.606 & 0.100
& 0.771 & 0.806 & 0.551 & 0.051
& 0.810 & 0.873 & 0.772 & 0.057 \\
MGL \cite{zhai2021mutual} \textsubscript{CVPR'21}
& \ding{55} & \ding{55} & \ding{55} 
& 0.893 & 0.923 & 0.813 & 0.030
& 0.775 & 0.847 & 0.673 & 0.088
& 0.814 & 0.865 & 0.666 & 0.035
& $\ddagger$ & $\ddagger$ & $\ddagger$ & $\ddagger$ \\
PFNet \cite{mei2021camouflaged} \textsubscript{CVPR'21}
& \ding{55} & \ding{55} & \ding{55} 
& 0.882 & 0.931 & 0.810 & 0.033
& 0.782 & 0.852 & 0.695 & 0.085
& 0.800 & 0.868 & 0.660 & 0.040
& 0.829 & 0.887 & 0.745 & 0.053 \\
UGTR \cite{yang2021uncertainty} \textsubscript{ICCV'21}
& \ding{55} & \ding{55} & \ding{55} 
& 0.888 & 0.940 & 0.794 & 0.031
& 0.785 & 0.859 & 0.686 & 0.086
& 0.818 & 0.850 & 0.667 & 0.035
& $\ddagger$ & $\ddagger$ & $\ddagger$ & $\ddagger$ \\
LSR \cite{lv2021simultaneously} \textsubscript{CVPR'21}
& \ding{55} & \ding{55} & \ding{55} 
& 0.893 & 0.938 & 0.839 & 0.033
& 0.793 & 0.826 & 0.725 & 0.085
& 0.793 & 0.868 & 0.685 & 0.041
& 0.839 & 0.883 & 0.779 & 0.053 \\
SINet-v2 \cite{fan2021concealed} \textsubscript{TPAMI'21}
& \ding{55} & \ding{55} & \ding{55} 
& 0.888 & 0.942 & 0.816 & 0.030
& 0.820 & 0.882 & 0.743 & 0.070
& 0.815 & 0.887 & 0.680 & 0.037
& 0.847 & 0.903 & 0.769 & 0.048 \\
PreyNet \textsuperscript{*} \cite{zhang2022preynet_mm} \textsubscript{MM'22}
& \ding{55} & \checkmark & \ding{55} 
& 0.895 & 0.951 & 0.844 & 0.028
& 0.790 & 0.842 & 0.708 & 0.077
& 0.813 & 0.891 & 0.697 & 0.034
& $\ddagger$ & $\ddagger$ & $\ddagger$ & $\ddagger$ \\
ZoomNet \textsuperscript{*} \cite{pang2022zoom} \textsubscript{CVPR'22}
& \ding{55} & \checkmark & \ding{55} 
& 0.902 & 0.958 & \underline{0.845} & \underline{0.023}
& \underline{0.820} & \underline{0.892} & \underline{0.752} & \textbf{0.066}
& 0.838 & 0.888 & 0.729 & \underline{0.029}
& 0.853 & 0.896 & 0.784 & 0.043 \\
PopNet \cite{wu-popNet} \textsubscript{ICCV'23} & \ding{55} & \checkmark & \ding{55} 
& \textbf{0.917} & \textbf{0.965} & \textbf{0.875} & \textbf{0.020}
& 0.808 & 0.859 & 0.744 & 0.077
& \textbf{0.851} & \textbf{0.910} & \textbf{0.757} & \textbf{0.028}
& \underline{0.861} & \underline{0.909} & \textbf{0.802} & 0.042 \\
FEDER \cite{he2023camouflaged} \textsubscript{CVPR'23}
& \ding{55} & \ding{55} & \ding{55} 
& \underline{0.907} & \underline{0.964} & $\ddagger$ & 0.025
& 0.807 & 0.873 & $\ddagger$ & 0.069
& 0.823 & 0.900 & $\ddagger$ & 0.032
& 0.846 & 0.905 & $\ddagger$ & 0.045 \\
RUN \cite{he2025run} \textsubscript{ICML'25}
& \ding{55} & \ding{55} & \ding{55} 
& 0.895 & 0.952 & $\ddagger$ & 0.027
& 0.806 & 0.868 & $\ddagger$ & 0.070
& 0.827 & \underline{0.903} & $\ddagger$ & 0.030
& 0.851 & 0.908 & $\ddagger$ & \underline{0.042} \\
\midrule
\rowcolor{gray!20}
Base-R (Ours) \textsuperscript{$\dagger$} & \ding{55} & \ding{55} & \checkmark 
& 0.880 & 0.949 & 0.840 & 0.025
& \textbf{0.827} & \textbf{0.898} & \textbf{0.765} & \underline{0.067}
& \underline{0.842} & 0.898 & \underline{0.744} & 0.030
& \textbf{0.865} & \textbf{0.915} & \underline{0.795} & \textbf{0.042} \\
\midrule
SAM \cite{kirillov2023segment} \textsubscript{ICCV'23} & \checkmark & \ding{55} & \ding{55} 
& $\ddagger$ & $\ddagger$ & $\ddagger$ & $\ddagger$
& $\ddagger$ & $\ddagger$ & $\ddagger$ & $\ddagger$
& 0.778 & 0.800 & 0.701 & 0.050
& 0.765 & 0.778 & 0.696 & 0.078 \\
HitNet \textsuperscript{*} \cite{hu2023high} \textsubscript{AAAI'23} & \ding{55} & \checkmark & \ding{55} 
& 0.922 & \textbf{0.970} & \textbf{0.903} & \textbf{0.018}
& 0.844 & 0.902 & 0.801 & 0.057
& 0.868 & 0.932 & 0.798 & 0.024
& 0.870 & 0.921 & 0.825 & 0.039 \\
FPNet \cite{cong2023frequency} \textsubscript{MM'23} & \ding{55} & \ding{55} & \ding{55} 
& 0.914 & 0.960 & 0.868 & 0.022
& 0.851 & 0.912 & 0.802 & 0.056
& 0.851 & 0.909 & 0.755 & 0.028
& $\ddagger$ & $\ddagger$ & $\ddagger$ & $\ddagger$ \\
FSPNet \cite{huang2023feature} \textsubscript{CVPR'23} & \ding{55} & \ding{55} & \ding{55} 
& 0.908 & 0.943 & 0.851 & 0.023
& 0.856 & 0.899 & 0.799 & 0.050
& 0.851 & 0.895 & 0.735 & 0.026
& 0.879 & 0.915 & 0.816 & 0.035 \\
EVP \cite{liu2023explicit} \textsubscript{CVPR'23}
& \ding{55} & \ding{55} & \ding{55} 
& 0.871 & 0.917 & 0.795 & 0.036
& 0.846 & 0.895 & 0.777 & 0.067
& 0.843 & 0.907 & 0.742 & 0.032
& 0.874 & $\ddagger$ & $\ddagger$ & $\ddagger$ \\
CamoFormer \cite{yin2024camoformer} \textsubscript{TPAMI'24} & \ding{55} & \ding{55} & \ding{55} 
& 0.910 & 0.957 & 0.866 & 0.022
& 0.872 & 0.929 & 0.831 & 0.046
& 0.869 & 0.932 & 0.786 & 0.023
& 0.892 & 0.939 & 0.847 & 0.030 \\
VSCode \cite{luo2024vscode} \textsubscript{CVPR'24}
& \ding{55} & \checkmark & \ding{55} 
& $\ddagger$ & $\ddagger$ & $\ddagger$ & $\ddagger$
& 0.836 & 0.892 & 0.768 & 0.060
& 0.847 & 0.913 & 0.744 & 0.028
& 0.874 & 0.920 & 0.813 & 0.038 \\
Spider \cite{zhao2024spider} \textsubscript{ICML'24} & \ding{55} & \ding{55} & \ding{55} 
& 0.906 & 0.951 & 0.848 & 0.025
& 0.855 & 0.908 & 0.799 & 0.053
& 0.856 & 0.917 & 0.756 & 0.028
& 0.880 & 0.925 & 0.825 & 0.036 \\
DSAM \cite{yu2024exploring} \textsubscript{MM'24}
& \checkmark & \checkmark & \ding{55} 
& 0.853 & 0.924 & 0.785 & 0.045
& 0.832 & 0.913 & 0.794 & 0.061
& 0.846 & 0.921 & 0.760 & 0.033
& 0.871 & 0.932 & 0.826 & 0.040 \\
CamoDiffusion-VAE \cite{sun2025conditional} \textsubscript{TPAMI'25}
& \ding{55} & \ding{55} & \ding{55}
& $\ddagger$ & $\ddagger$ & $\ddagger$ & $\ddagger$
& 0.868 & 0.926 & 0.827 & 0.047
& 0.857 & 0.919 & 0.759 & 0.024
& 0.877 & 0.926 & 0.821 & 0.034 \\
COMPrompter \cite{zhang2025comprompter} \textsubscript{SCIS'25}
& \checkmark & \ding{55} & \ding{55}
& 0.884 & 0.946 & 0.830 & 0.030
& 0.853 & 0.919 & 0.819 & 0.054
& 0.861 & 0.933 & 0.779 & 0.026
& 0.880 & 0.935 & 0.840 & 0.036 \\
USCNet \cite{zhou2025rethinking} \textsubscript{ICCV'25}
& \checkmark & \checkmark & \ding{55}
& $\ddagger$ & $\ddagger$ & $\ddagger$ & $\ddagger$
& 0.845 & 0.886 & 0.790 & 0.049
& 0.821 & 0.869 & 0.700 & 0.030
& 0.839 & 0.877 & 0.768 & 0.039 \\
VLSAM \cite{ren2025multi} \textsubscript{ICCV'25}
& \checkmark & \checkmark & \ding{55}
& \underline{0.923} & 0.946 & 0.853 & 0.027
& 0.863 & 0.901 & 0.782 & 0.059
& \textbf{0.896} & 0.907 & \textbf{0.808} & 0.023
& 0.872 & 0.918 & 0.819 & 0.033 \\
SAMTTT \cite{yu2025sam} \textsubscript{MM'25}
& \checkmark & \ding{55} & \checkmark
& $\ddagger$ & $\ddagger$ & $\ddagger$ & $\ddagger$
& 0.868 & 0.935 & 0.838 & 0.045
& 0.874 & \textbf{0.942} & \underline{0.805} & 0.027
& 0.884 & 0.943 & 0.837 & 0.031 \\
\midrule
\rowcolor{gray!20}
Base-P (Ours) \textsuperscript{$\dagger$}
& \ding{55} & \ding{55} & \checkmark
& 0.893 & 0.962 & 0.853 & 0.023
& \underline{0.873} & \textbf{0.940} & \underline{0.840} & \textbf{0.040}
& 0.860 & 0.933 & 0.781 & \underline{0.022}
& 0.891 & \textbf{0.950} & \underline{0.856} & \textbf{0.026} \\
\rowcolor{gray!20}
Spider \textsuperscript{$\dagger$} \cite{zhao2024spider}
& \ding{55} & \checkmark & \checkmark
& 0.919 & 0.963 & 0.868 & 0.022
& 0.869 & 0.918 & 0.815 & 0.049
& 0.872 & 0.929 & 0.774 & 0.025
& \underline{0.894} & 0.938 & 0.842 & 0.033 \\
\rowcolor{gray!20}
DSAM \textsuperscript{$\dagger$} \cite{yu2024exploring}
& \checkmark & \checkmark & \checkmark
& 0.868 & 0.937 & 0.807 & 0.041
& 0.847 & 0.926 & 0.814 & 0.056
& 0.861 & 0.934 & 0.779 & 0.030
& 0.885 & 0.942 & 0.839 & 0.037 \\
\rowcolor{gray!20}
CamoFormer \textsuperscript{$\dagger$} \cite{yin2024camoformer}
& \ding{55} & \ding{55} & \checkmark
& \textbf{0.923} & \underline{0.966} & \underline{0.883} & \underline{0.019}
& \textbf{0.885} & \underline{0.940} & \textbf{0.850} & \underline{0.042}
& \underline{0.884} & \underline{0.941} & 0.804 & \textbf{0.020}
& \textbf{0.907} & \underline{0.948} & \textbf{0.866} & \underline{0.027} \\
\bottomrule
\end{tabular}%
}
\label{tab:quantitative_comparison}
\end{table*}

\begin{table*}[tb]
\caption{Quantitative comparison on degraded benchmark datasets. Attr. indicates the applied degradation type: GN (Gaussian noise), GB (Gaussian blur), and CR (contrast reduction). Gray rows represent methods equipped with HCL. *: Depth maps generated via \cite{ranftl2021vision} following \cite{yu2024exploring,wu-popNet}.}
\centering
\resizebox{\linewidth}{!}{
\setlength\tabcolsep{4pt}
\begin{tabular}{l|c|cccc|cccc|cccc|cccc}
\toprule
\multirow{2}{*}{Method} 
& \multirow{2}{*}{Attr.} 
& \multicolumn{4}{c|}{CHAMELEON (76)}
& \multicolumn{4}{c|}{CAMO (250)}
& \multicolumn{4}{c|}{COD10K (2026)}
& \multicolumn{4}{c}{NC4K (4121)} \\
\cmidrule(lr){3-6} \cmidrule(lr){7-10} \cmidrule(lr){11-14} \cmidrule(lr){15-18}
& & $S_m \uparrow$ & $E_m \uparrow$ & $F^{w}_{\beta} \uparrow$ & MAE $\downarrow$
& $S_m \uparrow$ & $E_m \uparrow$ & $F^{w}_{\beta} \uparrow$ & MAE $\downarrow$
& $S_m \uparrow$ & $E_m \uparrow$ & $F^{w}_{\beta} \uparrow$ & MAE $\downarrow$
& $S_m \uparrow$ & $E_m \uparrow$ & $F^{w}_{\beta} \uparrow$ & MAE $\downarrow$ \\
\midrule
% --- GN Section ---
DSAM \textsuperscript{*} & & 0.805 & 0.894 & 0.714 & 0.059 & 0.768 & 0.863 & 0.704 & 0.087 & 0.808 & 0.897 & 0.700 & 0.043 & 0.828 & 0.907 & 0.765 & 0.054 \\
\rowcolor{gray!13} \quad + HCL & & 0.837 & 0.914 & 0.758 & 0.051 & 0.800 & 0.892 & 0.756 & 0.075 & 0.833 & 0.907 & 0.741 & 0.038 & 0.855 & 0.921 & 0.797 & 0.048 \\
Spider & & 0.877 & 0.927 & 0.800 & 0.034 & 0.765 & 0.828 & 0.668 & 0.089 & 0.810 & 0.881 & 0.686 & 0.038 & 0.831 & 0.888 & 0.754 & 0.052 \\
\rowcolor{gray!13} \quad + HCL & & 0.892 & 0.940 & 0.831 & 0.030 & 0.812 & 0.873 & 0.726 & 0.072 & 0.838 & 0.901 & 0.724 & 0.033 & 0.864 & 0.897 & 0.788 & 0.045 \\
CamoFormer & & 0.872 & 0.922 & 0.805 & 0.036 & 0.810 & 0.871 & 0.731 & 0.070 & 0.833 & 0.902 & 0.724 & 0.032 & 0.854 & 0.909 & 0.788 & 0.043 \\
\rowcolor{gray!13} \quad + HCL & \multirow{-2}{*}{GN} & 0.896 & 0.941 & 0.844 & 0.030 & 0.853 & 0.896 & 0.788 & 0.059 & 0.850 & 0.918 & 0.761 & 0.027 & 0.878 & 0.921 & 0.823 & 0.038 \\
SAM-TTT & & 0.848 & 0.922 & 0.798 & 0.042 & 0.832 & 0.910 & 0.792 & 0.058 & 0.842 & 0.918 & 0.778 & 0.034 & 0.864 & 0.928 & 0.808 & 0.040 \\
\rowcolor{gray!13} \quad + HCL & & 0.872 & 0.939 & 0.830 & 0.036 & 0.855 & 0.925 & 0.820 & 0.050 & 0.866 & 0.932 & 0.812 & 0.029 & 0.884 & 0.941 & 0.838 & 0.034 \\
Base (Ours) & & 0.802 & 0.887 & 0.740 & 0.043 & 0.753 & 0.840 & 0.709 & 0.075 & 0.785 & 0.868 & 0.686 & 0.038 & 0.807 & 0.875 & 0.760 & 0.042 \\
\rowcolor{gray!13} \quad + HCL & & 0.863 & 0.941 & 0.796 & 0.031 & 0.822 & 0.893 & 0.762 & 0.064 & 0.834 & 0.911 & 0.737 & 0.029 & 0.866 & 0.930 & 0.815 & 0.035 \\
\midrule \addlinespace[0.3em]

% --- GB Section ---
DSAM \textsuperscript{*} & & 0.813 & 0.896 & 0.731 & 0.059 & 0.794 & 0.883 & 0.738 & 0.082 & 0.813 & 0.896 & 0.707 & 0.042 & 0.839 & 0.913 & 0.779 & 0.051 \\
\rowcolor{gray!13} \quad + HCL & & 0.835 & 0.911 & 0.765 & 0.052 & 0.812 & 0.895 & 0.772 & 0.072 & 0.835 & 0.905 & 0.738 & 0.038 & 0.852 & 0.922 & 0.800 & 0.047 \\
Spider & & 0.866 & 0.922 & 0.785 & 0.039 & 0.759 & 0.808 & 0.647 & 0.091 & 0.802 & 0.869 & 0.668 & 0.044 & 0.824 & 0.878 & 0.739 & 0.057 \\
\rowcolor{gray!13} \quad + HCL & & 0.888 & 0.936 & 0.826 & 0.033 & 0.803 & 0.849 & 0.708 & 0.077 & 0.828 & 0.887 & 0.710 & 0.036 & 0.851 & 0.899 & 0.790 & 0.047 \\
CamoFormer & & 0.787 & 0.837 & 0.659 & 0.058 & 0.759 & 0.811 & 0.651 & 0.088 & 0.789 & 0.859 & 0.653 & 0.044 & 0.813 & 0.866 & 0.721 & 0.058 \\
\rowcolor{gray!13} \quad + HCL & \multirow{-2}{*}{GB} & 0.832 & 0.872 & 0.700 & 0.046 & 0.797 & 0.852 & 0.705 & 0.072 & 0.815 & 0.891 & 0.696 & 0.035 & 0.861 & 0.903 & 0.778 & 0.044 \\
SAM-TTT & & 0.838 & 0.915 & 0.782 & 0.046 & 0.825 & 0.902 & 0.776 & 0.061 & 0.835 & 0.910 & 0.760 & 0.037 & 0.855 & 0.920 & 0.795 & 0.042 \\
\rowcolor{gray!13} \quad + HCL & & 0.858 & 0.929 & 0.808 & 0.041 & 0.845 & 0.918 & 0.802 & 0.055 & 0.857 & 0.926 & 0.788 & 0.032 & 0.874 & 0.936 & 0.822 & 0.037 \\
Base (Ours) & & 0.786 & 0.874 & 0.725 & 0.047 & 0.752 & 0.835 & 0.699 & 0.078 & 0.771 & 0.843 & 0.645 & 0.048 & 0.795 & 0.862 & 0.744 & 0.046 \\
\rowcolor{gray!13} \quad + HCL & & 0.842 & 0.943 & 0.773 & 0.035 & 0.815 & 0.896 & 0.758 & 0.064 & 0.827 & 0.907 & 0.724 & 0.029 & 0.856 & 0.923 & 0.799 & 0.038 \\
\midrule \addlinespace[0.3em]

% --- CR Section ---
DSAM \textsuperscript{*} & & 0.846 & 0.923 & 0.773 & 0.046 & 0.828 & 0.908 & 0.787 & 0.064 & 0.840 & 0.917 & 0.750 & 0.034 & 0.868 & 0.930 & 0.822 & 0.041 \\
\rowcolor{gray!13} \quad + HCL & & 0.856 & 0.929 & 0.783 & 0.044 & 0.839 & 0.917 & 0.801 & 0.062 & 0.851 & 0.929 & 0.763 & 0.032 & 0.879 & 0.937 & 0.830 & 0.040 \\
Spider & & 0.905 & 0.951 & 0.845 & 0.025 & 0.850 & 0.904 & 0.790 & 0.055 & 0.850 & 0.914 & 0.746 & 0.029 & 0.877 & 0.923 & 0.820 & 0.037 \\
\rowcolor{gray!13} \quad + HCL & & 0.913 & 0.955 & 0.853 & 0.024 & 0.861 & 0.910 & 0.799 & 0.054 & 0.858 & 0.920 & 0.758 & 0.028 & 0.883 & 0.928 & 0.827 & 0.036 \\
CamoFormer & & 0.888 & 0.936 & 0.830 & 0.028 & 0.852 & 0.908 & 0.799 & 0.054 & 0.844 & 0.911 & 0.745 & 0.028 & 0.877 & 0.924 & 0.822 & 0.035 \\
\rowcolor{gray!13} \quad + HCL & \multirow{-2}{*}{CR} & 0.895 & 0.945 & 0.847 & 0.025 & 0.861 & 0.917 & 0.818 & 0.050 & 0.855 & 0.920 & 0.763 & 0.025 & 0.884 & 0.932 & 0.835 & 0.032 \\
SAM-TTT & & 0.882 & 0.938 & 0.830 & 0.031 & 0.858 & 0.925 & 0.815 & 0.049 & 0.865 & 0.930 & 0.792 & 0.029 & 0.876 & 0.936 & 0.825 & 0.035 \\
\rowcolor{gray!13} \quad + HCL & & 0.892 & 0.945 & 0.842 & 0.028 & 0.868 & 0.933 & 0.828 & 0.045 & 0.876 & 0.938 & 0.806 & 0.026 & 0.885 & 0.942 & 0.838 & 0.032 \\
Base (Ours) & & 0.822 & 0.901 & 0.754 & 0.041 & 0.825 & 0.888 & 0.775 & 0.056 & 0.802 & 0.874 & 0.715 & 0.033 & 0.830 & 0.889 & 0.783 & 0.037 \\
\rowcolor{gray!13} \quad + HCL & & 0.868 & 0.945 & 0.804 & 0.028 & 0.863 & 0.935 & 0.825 & 0.044 & 0.848 & 0.924 & 0.761 & 0.024 & 0.886 & 0.945 & 0.846 & 0.028 \\
\bottomrule
\end{tabular}%
}
\label{tab:Quantitative comparison on degraded COD benchmark datasets}
\end{table*}

\begin{table}[h]
\caption{Computational cost comparison. 
Values in parentheses denote the change after equipping HCL.}
\centering
% \small
\resizebox{\linewidth}{!}{%
\setlength{\tabcolsep}{2.5pt}
\renewcommand{\arraystretch}{1.15}
\begin{tabular}{l|c|c|c|c}
\toprule
\multirow{2}{*}{Method} 
& \multirow{2}{*}{Input Size} 
& \multicolumn{1}{c|}{Params (M)} 
& \multicolumn{1}{c|}{FLOPs (G)} 
& \multicolumn{1}{c}{FPS} \\
\cmidrule(lr){3-5}
& & {Baseline (+$\Delta$)} & {Baseline (+$\Delta$)} & {Baseline (-$\Delta$)} \\
\midrule
DSAM 
& 1024 × 1024 
& 121.93\, (+9.24) 
& 1132.15\, (+54.86) 
& 8\, (-1) \\
Spider 
& 384 × 384 
& 175.09\, (+9.24) 
& 48.59\, (+7.58) 
& 38\, (-7) \\
CamoFormer 
& 384 × 384
& 71.40\, (+9.24) 
& 50.12\, (+7.58) 
& 32\, (-6) \\
SAM-TTT 
& 1024 × 1024
& 101.80\, (+9.24) 
& 405.76\, (+54.86) 
& 12\, (-2) \\
Base (Ours) 
& 384 × 384 
& 40.58\, (+9.24) 
& 28.49\, (+7.58) 
& 42\, (-8) \\
\bottomrule
\end{tabular}
}
\label{tab:comp_cost}
\end{table}

\begin{table*}[tb]
\caption{Quantitative ablation of proposed components under different conditions on the COD10K (Top) and NC4K (Bottom) datasets. C1, C2, C3, C4, and C5 represent spatial reconstruction, low-frequency reconstruction, high-frequency reconstruction, confidence map-guided and variational fusion, respectively.}
\centering
\resizebox{\textwidth}{!}{
\setlength\tabcolsep{3pt}
\begin{tabular}{c|ccc|c|cc|cccc|cccc|cccc|cccc}
\toprule
\multirow{2}{*}{Variant} & \multicolumn{3}{c|}{HRR} & \multirow{2}{*}{TAG} & \multicolumn{2}{c|}{PCC}
& \multicolumn{4}{c|}{Clean}
& \multicolumn{4}{c|}{Noise}
& \multicolumn{4}{c|}{Blur}
& \multicolumn{4}{c}{Contrast} \\
\cmidrule(lr){2-4} \cmidrule(lr){6-7} \cmidrule(lr){8-11} \cmidrule(lr){12-15} \cmidrule(lr){16-19} \cmidrule(lr){20-23}
& C1 & C2 & C3 & & C4 & C5
& $S_m \uparrow$ & $E_m \uparrow$ & $F^{w}_{\beta} \uparrow$ & MAE $\downarrow$
& $S_m \uparrow$ & $E_m \uparrow$ & $F^{w}_{\beta} \uparrow$ & MAE $\downarrow$
& $S_m \uparrow$ & $E_m \uparrow$ & $F^{w}_{\beta} \uparrow$ & MAE $\downarrow$
& $S_m \uparrow$ & $E_m \uparrow$ & $F^{w}_{\beta} \uparrow$ & MAE $\downarrow$ \\
\midrule
I & & & & & & 
& 0.818 & 0.889 & 0.747 & 0.029
& 0.785 & 0.868 & 0.686 & 0.038
& 0.771 & 0.843 & 0.645 & 0.048
& 0.802 & 0.874 & 0.715 & 0.033 \\
II & \checkmark & & & & & 
& 0.827 & 0.898 & 0.755 & 0.027 
& 0.794 & 0.874 & 0.695 & 0.036 
& 0.780 & 0.855 & 0.657 & 0.044 
& 0.812 & 0.884 & 0.720 & 0.031  \\
III & & \checkmark & & & & 
& 0.822 & 0.894 & 0.753 & 0.028 
& 0.793 & 0.870 & 0.688 & 0.037
& 0.781 & 0.850 & 0.651 & 0.045
& 0.808 & 0.885 & 0.718 & 0.032 \\
IV & & & \checkmark & & & 
& 0.824 & 0.891 & 0.751 & 0.028 
& 0.789 & 0.873 & 0.690 & 0.036
& 0.778 & 0.853 & 0.648 & 0.043
& 0.805 & 0.880 & 0.721 & 0.032 \\
V & \checkmark & \checkmark & & & & 
& 0.830 & 0.903 & 0.759 & 0.027 
& 0.801 & 0.878 & 0.693 & 0.035
& 0.783 & 0.862 & 0.670 & 0.042
& 0.822 & 0.889 & 0.723 & 0.030 \\
VI & \checkmark & & \checkmark & & & 
& 0.831 & 0.902 & 0.754 & 0.026 
& 0.798 & 0.880 & 0.698 & 0.034
& 0.785 & 0.864 & 0.665 & 0.041
& 0.817 & 0.890 & 0.725 & 0.030 \\
VII & & \checkmark & \checkmark & & & 
& 0.826 & 0.899 & 0.755 & 0.025 
& 0.796 & 0.875 & 0.697 & 0.034 
& 0.784 & 0.860 & 0.669 & 0.040 
& 0.815 & 0.888 & 0.724 & 0.031  \\
VIII & \checkmark & \checkmark & \checkmark & & & 
& 0.839 & 0.907 & 0.761 & 0.024 
& 0.807 & 0.887 & 0.705 & 0.033 
& 0.793 & 0.875 & 0.684 & 0.037 
& 0.834 & 0.901 & 0.732 & 0.029  \\
IX & \checkmark & \checkmark & \checkmark & \checkmark & & 
& 0.844 & 0.914 & 0.765 & 0.024 
& 0.814 & 0.895 & 0.718 & 0.032 
& 0.803 & 0.888 & 0.700 & 0.035 
& 0.840 & 0.900 & 0.741 & 0.027  \\
X & \checkmark & \checkmark & \checkmark & \checkmark & \checkmark & 
& 0.848 & 0.920 & 0.771 & 0.023 
& 0.820 & 0.902 & 0.725 & 0.031
& 0.810 & 0.896 & 0.711 & 0.032 
& 0.839 & 0.911 & 0.748 & 0.026  \\
\rowcolor{gray!20}
XI & \checkmark & \checkmark & \checkmark & \checkmark & \checkmark & \checkmark
& \textbf{0.860} & \textbf{0.933} & \textbf{0.781} & \textbf{0.022}
& \textbf{0.834} & \textbf{0.911} & \textbf{0.737} & \textbf{0.029}
& \textbf{0.827} & \textbf{0.907} & \textbf{0.724} & \textbf{0.029}
& \textbf{0.848} & \textbf{0.924} & \textbf{0.761} & \textbf{0.024} \\
\midrule
XII & & & & & & 
& 0.834 & 0.900 & 0.801 & 0.032
& 0.807 & 0.875 & 0.760 & 0.042
& 0.795 & 0.862 & 0.744 & 0.046
& 0.830 & 0.889 & 0.783 & 0.037 \\
XIII & \checkmark & & & & & 
& 0.843 & 0.914 & 0.811 & 0.031
& 0.819 & 0.886 & 0.771 & 0.041
& 0.811 & 0.874 & 0.756 & 0.044
& 0.843 & 0.900 & 0.795 & 0.037 \\
XIV & & \checkmark & & & & 
& 0.838 & 0.907 & 0.809 & 0.032
& 0.815 & 0.880 & 0.766 & 0.042
& 0.803 & 0.868 & 0.749 & 0.045
& 0.835 & 0.897 & 0.788 & 0.037 \\
XV & & & \checkmark & & & 
& 0.840 & 0.909 & 0.805 & 0.031
& 0.810 & 0.883 & 0.764 & 0.042
& 0.804 & 0.869 & 0.751 & 0.045
& 0.837 & 0.893 & 0.790 & 0.038 \\
XVI & \checkmark & \checkmark & & & & 
& 0.847 & 0.919 & 0.820 & 0.030
& 0.826 & 0.894 & 0.776 & 0.040
& 0.816 & 0.882 & 0.761 & 0.043
& 0.851 & 0.905 & 0.807 & 0.035 \\
XVII & \checkmark & & \checkmark & & & 
& 0.845 & 0.921 & 0.815 & 0.030
& 0.824 & 0.897 & 0.774 & 0.041
& 0.820 & 0.877 & 0.763 & 0.044
& 0.853 & 0.907 & 0.803 & 0.034 \\
XVIII & & \checkmark & \checkmark & & & 
& 0.842 & 0.912 & 0.817 & 0.030
& 0.820 & 0.885 & 0.772 & 0.040
& 0.809 & 0.876 & 0.759 & 0.042
& 0.841 & 0.902 & 0.792 & 0.036 \\
XIX & \checkmark & \checkmark & \checkmark & & & 
& 0.859 & 0.930 & 0.835 & 0.028
& 0.845 & 0.909 & 0.784 & 0.039
& 0.829 & 0.894 & 0.771 & 0.042
& 0.870 & 0.920 & 0.828 & 0.031 \\
XX & \checkmark & \checkmark & \checkmark & \checkmark & & 
& 0.870 & 0.935 & 0.838 & 0.027
& 0.842 & 0.918 & 0.796 & 0.038
& 0.837 & 0.905 & 0.783 & 0.041
& 0.876 & 0.925 & 0.837 & 0.029 \\
XXI & \checkmark & \checkmark & \checkmark & \checkmark & \checkmark & 
& 0.878 & 0.941 & 0.844 & 0.026
& 0.852 & 0.923 & 0.802 & 0.036
& 0.843 & 0.913 & 0.787 & 0.040
& 0.880 & 0.931 & 0.833 & 0.028 \\
\rowcolor{gray!20}
XXII & \checkmark & \checkmark & \checkmark & \checkmark & \checkmark & \checkmark
& \textbf{0.891} & \textbf{0.950} & \textbf{0.856} & \textbf{0.026}
& \textbf{0.866} & \textbf{0.930} & \textbf{0.815} & \textbf{0.035}
& \textbf{0.856} & \textbf{0.923} & \textbf{0.799} & \textbf{0.038}
& \textbf{0.886} & \textbf{0.945} & \textbf{0.846} & \textbf{0.028} \\
\bottomrule
\end{tabular}%
}
\label{tab:ablation_study}
\end{table*}

\begin{figure*}[tb]
  \centering
  \includegraphics[width=1\textwidth]{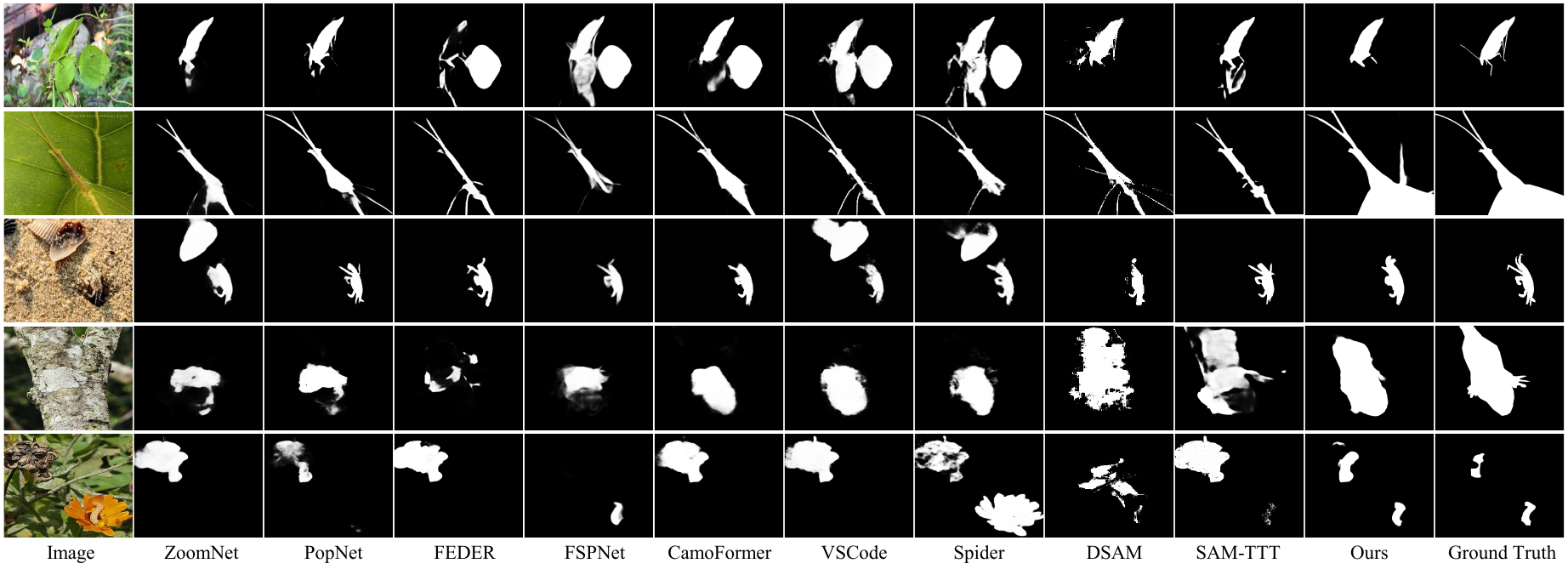}
  \caption{Qualitative comparison on normal scenarios. Best viewed by zooming in.}
  \label{fig:COD_Vis_Comparision}
\end{figure*}

\begin{figure*}[tb]
  \centering
\includegraphics[width=1\textwidth]{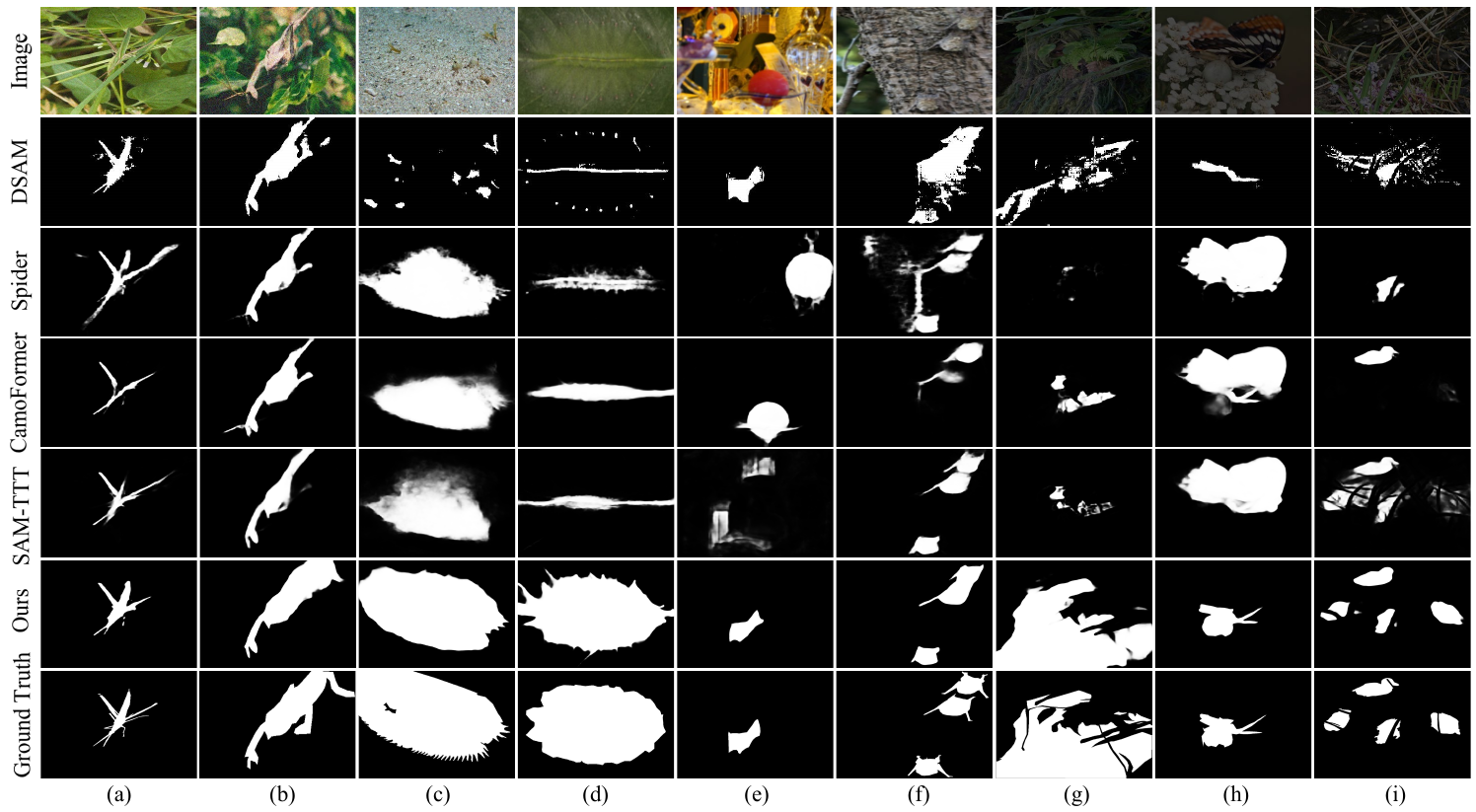}
  \caption{Qualitative comparison on degraded camouflaged scenarios. (a)-(c): GN setting, (d)-(f): GB setting, (g)-(i): CR setting.}
\label{fig:COD_Domain_Vis_Comparision}
\end{figure*}

\section{EXPERIMENTS}

\subsection{Datasets}
We conduct experiments on four standard COD benchmark datasets. CAMO \cite{le2019anabranch} and COD10K \cite{fan2020camouflaged} contain 1,000 and 3,040 training pairs, with 250 and 2,026 testing pairs, respectively. CHAMELEON \cite{skurowski2018animal} and NC4K \cite{lv2021simultaneously} are used exclusively for testing, consisting of 76 and 4,121 images. Following \cite{hao2025simple,cong2023frequency}, we aggregate the training pairs from CAMO and COD10K as the full training set, while the remaining data is assigned as the testing set. For the underwater benchmarks, following \cite{zha2025heterogeneous}, we select MAS3K \cite{li2020mas3k}, RMAS \cite{fu2023masnet}, UFO120 \cite{islam2020simultaneous}, and RUWI \cite{drews2021underwater}. MAS3K consists of 1,769 training pairs and 1,141 testing pairs with foreground objects. RMAS and UFO120 contain 3,014/500 and 1,500/120 training/testing pairs, respectively. Following \cite{zha2025heterogeneous}, we set the training and testing sets of RUWI to 525 and 175 pairs, respectively. To further evaluate the model’s robustness under distribution shifts, inspired by \cite{hendrycks2019benchmarking,huang2024learning}, we adopt three common degradation strategies: Gaussian blur (GB), Gaussian noise (GN), and contrast reduction (CR). These perturbations simulate real-world variations and assess the model’s adaptability in challenging conditions.

% toolbox image and prompt encode
\subsection{Implementation Details.} 
We implement our model and conduct experiments using the PyTorch framework on NVIDIA A100 GPUs. We utilize PVT-v2 \cite{wang2022pvt} and ResNet50 \cite{he2016deep}, both pre-trained on ImageNet, as backbone networks. For fair comparison, following \cite{he2023camouflaged,huang2023feature}, we set the input resolution to 384×384. During the training phase, we leverage the AdamW optimizer with a batch size of 16, an initial learning rate of 0.001, and train for 200 epochs. Data augmentation strategies, including horizontal and vertical flipping, are applied to enhance generalization. During the testing phase, we apply the same settings for the TTA stage as in the training phase. Moreover, we do not incorporate any post-processing techniques, such as CRF, to refine predictions. For the reconstruction branch, we set the patch size to 16, the depth of the encoder-decoder and the number of attention heads to 4, with the embedding dimension of 512 to reduce computational cost. To ensure a fair comparison, we compute the evaluation metrics using either the prediction results, pre-trained weights, or source codes provided by the projects.

\subsection{Evaluation Metrics.} 

We adopt four evaluation metrics: S-measure ($S_m$), mean E-measure ($E_m$), weighted F-measure ($F^{w}_{\beta}$), and Mean Absolute Error (MAE). Note that for the first three metrics, higher values indicate better performance, whereas for MAE, lower values are preferable.

\subsection{Comparison with State-of-the-Art Methods}
\textit{1) Quantitative Comparison.} In Table \ref{tab:quantitative_comparison}, without relying on foundation models and incorporating depth maps, our method outperforms DSAM by 2.0\%, 1.8\%, 3.0\%, and 1.4\% across four metrics on the most challenging NC4K dataset. We analyze that SAM, though pre-trained on large-scale natural images, tends to exhibit biases in camouflaged scenarios and struggles to adapt effectively with limited parameter updates and prior prompts. Moreover, our method also surpasses HitNet by 2.1\%, 2.9\%, 3.1\%, and 1.3\%. We argue that although HitNet improves representation capacity by increasing input resolution to preserve discriminative features, it lacks sensitivity to the diversity of camouflage patterns, thereby limiting the adaptability in complex settings. In Table \ref{tab:Quantitative comparison on degraded COD benchmark datasets}, under three interference conditions, CR shows the least performance degradation, followed by GN, with GB exhibiting the greatest decline. From the frequency perspective, CR preserves low-frequency structures essential for shape perception, GN adds high-frequency noise that can be partially suppressed, whereas GB removes critical high-frequency details, directly impairing boundary localization. Our method demonstrates consistent improvement of approximately 2\%-8\% on the NC4K. The higher gains in degraded scenes highlight the effectiveness of representation consistency and self-calibration. We integrate HCL into other methods and observe substantial performance improvements, indicating that train-test distribution shifts inherently exist and require additional customized components. In Table \ref{tab:comp_cost}, we report the parameter count, computational cost (FLOPs), and inference speed (FPS) of the model after integrating HCL, measured on an A100 GPU. While HCL involves multiple auxiliary branches during the adaptation phase (such as spatial and frequency reconstruction), the computational increase remains manageable. For standard resolution inputs (384×384), integrating HCL only adds 9.24M parameters and 7.58G FLOPs. For our base model (Base-P), the inference speed drops slightly from 42 to 34 FPS, which still meets real-time processing requirements. For high-resolution inputs (1024×1024), when applied to models like DSAM and SAM-TTT, the parameter overhead remains constant at +9.24M. While the FLOPs naturally scale with the resolution (an increase of 54.86G), the impact on actual inference speed is minimal, experiencing only a negligible drop of 1 to 2 FPS (\textit{e.g.,} DSAM drops from 8 to 7 FPS, and SAM-TTT drops from 12 to 10 FPS). The substantial robustness and generalization gains achieved under severe distribution shifts effectively justify this modest computational trade-off. 

\textit{2) Qualitative Comparison.} In Figure~\ref{fig:COD_Vis_Comparision} and Figure~\ref{fig:COD_Domain_Vis_Comparision}, we present visualization results across various scenarios. Existing methods are prone to false negatives and false positives due to intrinsic physical characteristics and external disturbances, particularly in cases involving significant scale variations and multiple targets. In contrast, our approach effectively mitigates background clutter and imaging artifacts, accurately and comprehensively locating camouflaged regions.  

\begin{figure*}[tb]
  \centering
  \includegraphics[width=1\textwidth]{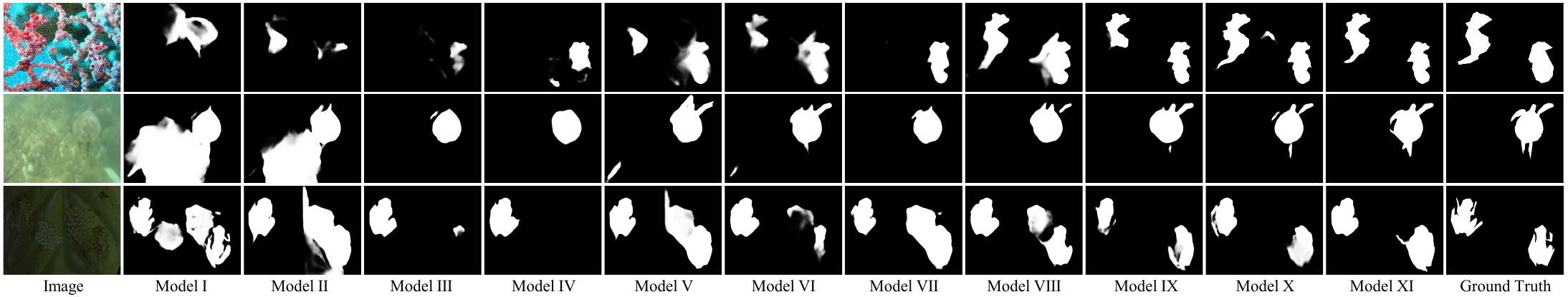}
  \caption{Qualitative ablation of the proposed components. Variants corresponding to Table \ref{tab:ablation_study}.}
  \label{fig:Ab_vis}
\end{figure*}

\begin{figure*}[tb]
  \centering
  \includegraphics[width=1\textwidth]{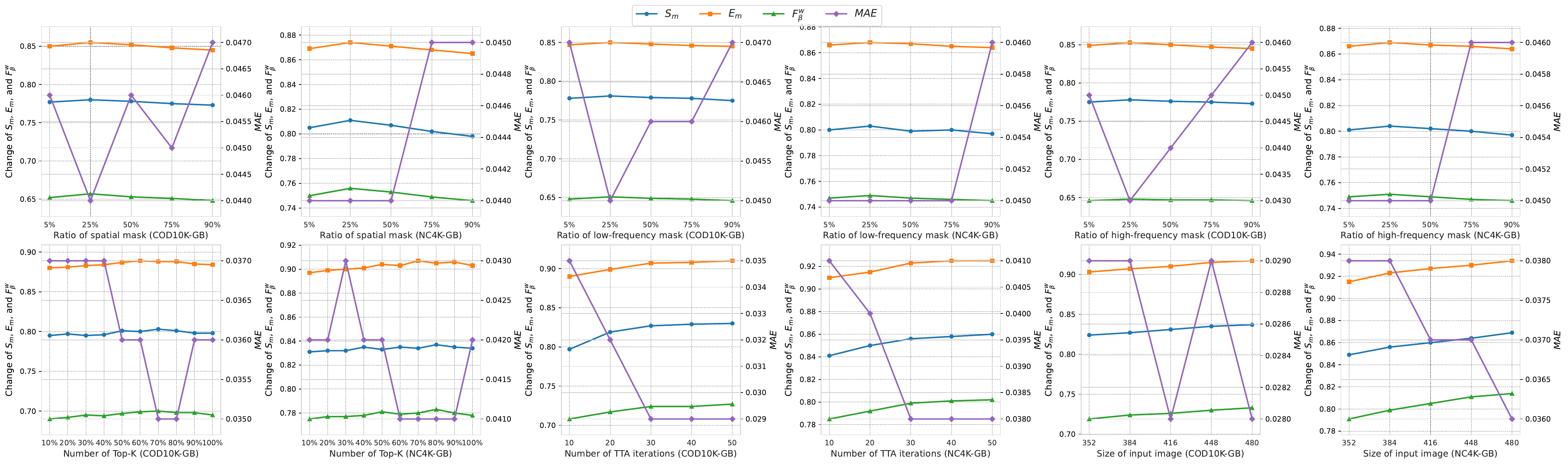}
  \caption{Quantitative ablation on the ratios of spatial, low-frequency, and high-frequency masks (with optimal spatial ratio), as well as the effects of Top-K selection, TTA iterations, and input image resolution}
  \label{fig:Hyperparameter settings}
\end{figure*}

\subsection{Ablation Study}
We quantitatively and qualitatively validate the impact of the proposed components, method variants, and critical hyperparameter settings.

\textit{1) Effect of the crucial components.} In Table \ref{tab:ablation_study}, we integrate components through one-step process. We find that: 1) For the HRR, directly using spatial reconstruction (\textit{i.e.,} vanilla MAE) does not yield significant results, while combining with frequency reconstruction achieves complementary effects, especially in degraded scenarios. 2) Integrating the TAG generally improves performance, although some cases show a decline. 3) The confidence map guidance and variational fusion components exhibit progressive coupling. We analyze: 1) Spatial and frequency domains emphasize different aspects of information recovery, and the variations in dynamic scenes and camouflage patterns necessitate more comprehensive clue extraction. 2) We utilize the TAG as a bridge to inject prior knowledge, enhancing detection representation when reconstruction results are ideal; conversely, it may weaken or even obscure when results are suboptimal. 3) The confidence map provides the robust prototype, while the variational component addresses internal instabilities and weight biases, transforming point estimates into probability distributions to mitigate the turbulence. In Figure \ref{fig:Ab_vis}, as the proposed components are gradually integrated, the detection areas approach the ground truth. In Table \ref{tab:Quantitative comparison on degraded COD benchmark datasets}, embedding the HCL into other methods can also effectively boost performance by about 3\%. The unified principle driving the HCL framework is the maximization of the Evidence Lower Bound (ELBO) \cite{Kingma2014} for the joint distribution of camouflaged object detection and structural reconstruction during test-time, which treats adaptation as an optimization of the posterior distribution of the latent representation conditioned on the observed image. To achieve this under severe distribution shifts, each component plays a theoretically necessary and complementary role: first, the HRR acts as an unbiased structural anchor that minimizes the uncertainty of the latent manifold by enforcing physical consistency across spatial and spectral domains; second, the TAG ensures task isomorphism by explicitly aligning the discriminative feature manifold with the structural reconstruction manifold to mitigate semantic drift; and finally, the PCC functions as a probabilistic filter that minimizes prediction risk by mapping deterministic prototypes into a probabilistic latent space to suppress high-variance noise. Together, these three components systematically prevent the model from collapsing into trivial solutions and ensure robust feature recalibration.

\begin{table}[tb] 
\caption{Quantitative comparison on degraded benchmark datasets under different adaptation strategies.}
\centering
\resizebox{\linewidth}{!}{
\setlength\tabcolsep{2.5pt}
\begin{tabular}{l|cccc|cccc}
\toprule
\multirow{2}{*}{Method} 
& \multicolumn{4}{c|}{COD10K-GB}
& \multicolumn{4}{c}{NC4K-GB} \\
\cmidrule(lr){2-5} \cmidrule(lr){6-9}
& $S_m \uparrow$ & $E_m \uparrow$ & $F^{w}_{\beta} \uparrow$ & MAE $\downarrow$
& $S_m \uparrow$ & $E_m \uparrow$ & $F^{w}_{\beta} \uparrow$ & MAE $\downarrow$ \\
\midrule
DSAM \cite{yu2024exploring} 
& 0.813 & 0.896 & 0.707 & 0.042 & 0.839 & 0.913 & 0.779 & 0.051 \\
\quad + TTT-Rot \cite{sun2020test} & 0.820 & 0.900 & 0.721 & 0.040 & 0.845 & 0.917 & 0.788 & 0.049 \\
\quad + TENT \cite{wang2020tent} & 0.817 & 0.898 & 0.724 & 0.039 & 0.842 & 0.915 & 0.784 & 0.049 \\
\quad + UDA \cite{mirza2022norm} & 0.818 & 0.899 & 0.718 & 0.040 & 0.843 & 0.916 & 0.782 & 0.050 \\
\quad + GH \cite{huang2024gradient} & 0.823 & 0.900 & 0.716 & 0.040 & 0.845 & 0.915 & 0.785 & 0.049 \\
\quad + TTT-MIM \cite{mansour2024ttt} & 0.822 & 0.902 & 0.726 & 0.039 & 0.847 & 0.919 & 0.790 & 0.048 \\
\rowcolor{gray!13} \quad + HCL & 0.835 & 0.905 & 0.738 & 0.038 & 0.852 & 0.922 & 0.800 & 0.047 \\
\midrule \addlinespace[0.3em]

Spider \cite{zhao2024spider} 
& 0.802 & 0.869 & 0.668 & 0.044 & 0.824 & 0.878 & 0.739 & 0.057 \\
\quad + TTT-Rot \cite{sun2020test} & 0.810 & 0.876 & 0.682 & 0.042 & 0.832 & 0.885 & 0.748 & 0.054 \\
\quad + TENT \cite{wang2020tent} & 0.806 & 0.872 & 0.690 & 0.041 & 0.829 & 0.881 & 0.744 & 0.054 \\
\quad + UDA \cite{mirza2022norm} & 0.807 & 0.873 & 0.679 & 0.042 & 0.830 & 0.882 & 0.742 & 0.055 \\
\quad + GH \cite{huang2024gradient} & 0.808 & 0.874 & 0.675 & 0.043 & 0.828 & 0.885 & 0.745 & 0.055 \\
\quad + TTT-MIM \cite{mansour2024ttt} & 0.812 & 0.879 & 0.694 & 0.041 & 0.835 & 0.888 & 0.752 & 0.053 \\
\rowcolor{gray!13} \quad + HCL & 0.828 & 0.887 & 0.710 & 0.036 & 0.851 & 0.899 & 0.790 & 0.047 \\
\midrule \addlinespace[0.3em]

CamoFormer \cite{yin2024camoformer}
& 0.789 & 0.859 & 0.653 & 0.044 & 0.813 & 0.866 & 0.721 & 0.058 \\
\quad + TTT-Rot \cite{sun2020test} & 0.798 & 0.869 & 0.668 & 0.042 & 0.822 & 0.874 & 0.735 & 0.055 \\
\quad + TENT \cite{wang2020tent} & 0.794 & 0.863 & 0.676 & 0.041 & 0.818 & 0.870 & 0.731 & 0.055 \\
\quad + UDA \cite{mirza2022norm} & 0.796 & 0.865 & 0.664 & 0.042 & 0.820 & 0.872 & 0.728 & 0.056 \\
\quad + GH \cite{huang2024gradient} & 0.795 & 0.869 & 0.677 & 0.042 & 0.832 & 0.875 & 0.733 & 0.055 \\
\quad + TTT-MIM \cite{mansour2024ttt} & 0.801 & 0.872 & 0.681 & 0.040 & 0.826 & 0.879 & 0.742 & 0.053 \\
\rowcolor{gray!13} \quad + HCL & 0.815 & 0.891 & 0.696 & 0.035 & 0.861 & 0.903 & 0.778 & 0.044 \\
\midrule \addlinespace[0.3em]

Base (Ours)
& 0.771 & 0.843 & 0.645 & 0.048 & 0.795 & 0.862 & 0.744 & 0.046 \\
\quad + TTT-Rot \cite{sun2020test} & 0.795 & 0.874 & 0.694 & 0.033 & 0.827 & 0.875 & 0.764 & 0.045 \\
\quad + TENT \cite{wang2020tent} & 0.782 & 0.845 & 0.700 & 0.035 & 0.816 & 0.867 & 0.752 & 0.041 \\
\quad + UDA \cite{mirza2022norm} & 0.800 & 0.866 & 0.687 & 0.036 & 0.824 & 0.879 & 0.748 & 0.044 \\
\quad + GH \cite{huang2024gradient} & 0.795 & 0.858 & 0.691 & 0.037 & 0.811 & 0.882 & 0.759 & 0.043 \\
\quad + TTT-MIM \cite{mansour2024ttt} & 0.792 & 0.879 & 0.695 & 0.034 & 0.821 & 0.890 & 0.762 & 0.043 \\
\rowcolor{gray!13} \quad + HCL & 0.827 & 0.907 & 0.724 & 0.029 & 0.856 & 0.923 & 0.799 & 0.038 \\
\bottomrule
\end{tabular}
}
\label{tab:TTA_methods_comparison}
\end{table}

\begin{table}[t]
\centering
\caption{Quantitative comparison of mask strategies.}
\resizebox{\linewidth}{!}{% Adjust the table to fit the line width
\setlength\tabcolsep{2.5pt}
\begin{tabular}{c|cccc|cccc}
\toprule
Method & \multicolumn{4}{c|}{COD10K-GB} & \multicolumn{4}{c}{NC4K-GB} \\
\midrule
& $S_m \uparrow$ & $E_m \uparrow$ & $F^{w}_{\beta} \uparrow$ & MAE $\downarrow$ 
& $S_m \uparrow$ & $E_m \uparrow$ & $F^{w}_{\beta} \uparrow$ & MAE $\downarrow$ \\
\midrule
Round & 0.820 & \textbf{0.909} & 0.713 & 0.030 & 0.849 & \textbf{0.925} & 0.789 & 0.039 \\
Grid & \textbf{0.829} & 0.900 & 0.715 & \textbf{0.028} & \underline{0.853} & 0.920 & 0.791 & \underline{0.039} \\
Block & 0.818 & 0.904 & \underline{0.718} & 0.029 & 0.848 & 0.918 & \underline{0.792} & 0.038 \\
\rowcolor{gray!20}
Random & \underline{0.827} & \underline{0.907} & \textbf{0.724} & \underline{0.029} & \textbf{0.856} & \underline{0.923} & \textbf{0.799} & \textbf{0.038} \\
\bottomrule
\end{tabular}
}
\label{tab:mask strategies}
\end{table}

\begin{table}[t]
\centering
\caption{Quantitative comparison of interactive strategies.}
\resizebox{\linewidth}{!}{% Adjust the table to fit the line width
\setlength\tabcolsep{2.5pt}
\begin{tabular}{c|cccc|cccc}
\toprule
Method & \multicolumn{4}{c|}{COD10K-GB} & \multicolumn{4}{c}{NC4K-GB} \\
\midrule
& $S_m \uparrow$ & $E_m \uparrow$ & $F^{w}_{\beta} \uparrow$ & MAE $\downarrow$ 
& $S_m \uparrow$ & $E_m \uparrow$ & $F^{w}_{\beta} \uparrow$ & MAE $\downarrow$ \\
\midrule
Spatial Cross & 0.797 & 0.880 & \underline{0.693} & 0.036 & \textbf{0.840} & 0.898 & 0.774 & 0.042 \\
Channel Cross & 0.795 & \textbf{0.889} & 0.690 & 0.037 & 0.833 & \underline{0.900} & \underline{0.780} & \underline{0.042} \\
TAG & \textbf{0.803} & \underline{0.888} & \textbf{0.700} & \textbf{0.035} & \underline{0.837} & \textbf{0.905} & \textbf{0.783} & \textbf{0.041} \\
\bottomrule
\end{tabular}
}
\label{tab:interactive strategy}
\end{table}

\begin{table}[t]
\centering
\caption{Quantitative comparison of prototype integration strategies.}
\resizebox{\linewidth}{!}{% Adjust the table to fit the line width
\setlength\tabcolsep{2.5pt}
\begin{tabular}{c|cccc|cccc}
\toprule
Method & \multicolumn{4}{c|}{COD10K-GB} & \multicolumn{4}{c}{NC4K-GB} \\
\midrule
& $S_m \uparrow$ & $E_m \uparrow$ & $F^{w}_{\beta} \uparrow$ & MAE $\downarrow$ 
& $S_m \uparrow$ & $E_m \uparrow$ & $F^{w}_{\beta} \uparrow$ & MAE $\downarrow$ \\
\midrule
Sum & \underline{0.823} & 0.901 & 0.717 & 0.032 & \underline{0.848} & 0.914 & \underline{0.795} & 0.040 \\
Concat & 0.815 & 0.899 & \underline{0.718} & 0.032 & 0.840 & \underline{0.917} & 0.790 & 0.039 \\
Cross Attention & 0.820 & \underline{0.902} & 0.715 & \underline{0.031} & 0.847 & 0.915 & 0.792 & \underline{0.039} \\
\rowcolor{gray!20}
Variational Fusion & \textbf{0.827} & \textbf{0.907} & \textbf{0.724} & \textbf{0.029} & \textbf{0.856} & \textbf{0.923} & \textbf{0.799} & \textbf{0.038} \\
\bottomrule
\end{tabular}
}
\label{tab:prototype integration strategy}
\end{table}

\begin{table*}[tb]
\caption{Quantitative comparison on degraded underwater benchmark datasets. Attr. indicates the applied degradation type: GN (Gaussian noise), GB (Gaussian blur), and CR (contrast reduction). N/A indicates no degradation. Gray rows represent methods equipped with HCL.}
\centering
\resizebox{\linewidth}{!}{
\setlength\tabcolsep{4pt}
\begin{tabular}{l|c|cccc|cccc|cccc|cccc}
\toprule
\multirow{2}{*}{Method} 
& \multirow{2}{*}{Attr.} 
& \multicolumn{4}{c|}{MAS3K (1141)}
& \multicolumn{4}{c|}{RMAS (500)}
& \multicolumn{4}{c|}{UFO120 (120)}
& \multicolumn{4}{c}{RUWI (175)} \\
\cmidrule(lr){3-6} \cmidrule(lr){7-10} \cmidrule(lr){11-14} \cmidrule(lr){15-18}
& & $S_m \uparrow$ & $E_m \uparrow$ & $F^{w}_{\beta} \uparrow$ & MAE $\downarrow$
& $S_m \uparrow$ & $E_m \uparrow$ & $F^{w}_{\beta} \uparrow$ & MAE $\downarrow$
& $S_m \uparrow$ & $E_m \uparrow$ & $F^{w}_{\beta} \uparrow$ & MAE $\downarrow$
& $S_m \uparrow$ & $E_m \uparrow$ & $F^{w}_{\beta} \uparrow$ & MAE $\downarrow$ \\
\midrule
DualSAM \cite{zhang2024fantastic} 
& \multirow{2}{*}{N/A} 
& 0.884 & 0.933 & 0.838 & 0.023 & 0.860 & 0.944 & 0.812 & 0.022 & 0.856 & 0.914 & 0.864 & 0.064 & 0.903 & 0.959 & 0.939 & 0.035 \\
MASSAM \cite{yan2024mas} 
& & 0.887 & 0.938 & 0.840 & 0.025 & 0.865 & 0.948 & 0.819 & 0.021 & 0.861 & 0.914 & 0.864 & 0.063 & 0.894 & 0.961 & 0.941 & 0.035 \\
\midrule \addlinespace[0.3em]

% --- GN Section ---
DualSAM \cite{zhang2024fantastic} & & 0.814 & 0.870 & 0.755 & 0.045 & 0.749 & 0.781 & 0.613 & 0.041 & 0.798 & 0.865 & 0.793 & 0.085 & 0.783 & 0.872 & 0.827 & 0.056 \\
\rowcolor{gray!13} \quad + HCL & & 0.853 & 0.901 & 0.799 & 0.032 & 0.785 & 0.833 & 0.651 & 0.033 & 0.827 & 0.892 & 0.831 & 0.075 & 0.832 & 0.903 & 0.869 & 0.045 \\
MASSAM \cite{yan2024mas} & \multirow{-2}{*}{GN} & 0.833 & 0.866 & 0.748 & 0.042 & 0.733 & 0.766 & 0.587 & 0.047 & 0.784 & 0.850 & 0.779 & 0.083 & 0.773 & 0.853 & 0.812 & 0.053 \\
\rowcolor{gray!13} \quad + HCL & & 0.844 & 0.900 & 0.777 & 0.036 & 0.770 & 0.801 & 0.629 & 0.038 & 0.814 & 0.870 & 0.794 & 0.077 & 0.837 & 0.898 & 0.867 & 0.048 \\
\midrule \addlinespace[0.3em]

% --- GB Section ---
DualSAM \cite{zhang2024fantastic} & & 0.795 & 0.857 & 0.746 & 0.050 & 0.740 & 0.769 & 0.615 & 0.044 & 0.778 & 0.848 & 0.758 & 0.082 & 0.795 & 0.863 & 0.834 & 0.051 \\
\rowcolor{gray!13} \quad + HCL & & 0.844 & 0.895 & 0.788 & 0.037 & 0.784 & 0.808 & 0.648 & 0.035 & 0.811 & 0.875 & 0.799 & 0.073 & 0.838 & 0.912 & 0.873 & 0.045 \\
MASSAM \cite{yan2024mas} & \multirow{-2}{*}{GB} & 0.785 & 0.844 & 0.750 & 0.046 & 0.745 & 0.781 & 0.620 & 0.043 & 0.773 & 0.835 & 0.739 & 0.079 & 0.763 & 0.849 & 0.830 & 0.055 \\
\rowcolor{gray!13} \quad + HCL & & 0.839 & 0.875 & 0.770 & 0.037 & 0.769 & 0.800 & 0.649 & 0.037 & 0.810 & 0.874 & 0.786 & 0.071 & 0.803 & 0.891 & 0.878 & 0.047 \\
\midrule \addlinespace[0.3em]

% --- CR Section ---
DualSAM \cite{zhang2024fantastic} & & 0.844 & 0.898 & 0.807 & 0.035 & 0.833 & 0.909 & 0.778 & 0.029 & 0.827 & 0.876 & 0.820 & 0.074 & 0.873 & 0.941 & 0.918 & 0.042 \\
\rowcolor{gray!13} \quad + HCL & & 0.867 & 0.921 & 0.819 & 0.031 & 0.840 & 0.930 & 0.790 & 0.026 & 0.844 & 0.899 & 0.840 & 0.071 & 0.888 & 0.945 & 0.938 & 0.038 \\
MASSAM \cite{yan2024mas} & \multirow{-2}{*}{CR} & 0.849 & 0.908 & 0.803 & 0.036 & 0.828 & 0.911 & 0.772 & 0.031 & 0.820 & 0.869 & 0.823 & 0.073 & 0.865 & 0.935 & 0.911 & 0.044 \\
\rowcolor{gray!13} \quad + HCL & & 0.864 & 0.916 & 0.824 & 0.032 & 0.844 & 0.924 & 0.796 & 0.027 & 0.834 & 0.883 & 0.833 & 0.069 & 0.882 & 0.940 & 0.924 & 0.038 \\
\bottomrule
\end{tabular}%
}
\label{tab:Quantitative comparison on degraded underwater benchmark datasets}
\end{table*}

\textit{2) Analysis of TTA strategies.} In Table \ref{tab:TTA_methods_comparison}, we consider: 1) Dynamically adjusting the statistics or parameters of the Batch Normalization layer to adapt to the distribution of the target domain \cite{wang2020tent,mirza2022norm}; 2) Updating parameters through rotation predictions to construct consistency \cite{sun2020test}. Our HCL shows the best performance. We argue that: 1) Both TENT and UDA struggle to handle confusable visual space and structural ambiguity; 2) Due to the high coupling between foreground and background, camouflaged regions can be positioned anywhere while preserving characteristics. In other words, rotation is inadequate for capturing camouflage and degradation cues. In contrast, image reconstruction explores the intrinsic relationships between regions, and the representations learned through self-supervised learning align better with the detection branch. Mainstream TTA methods (\textit{e.g.,} TENT) are primarily confidence-driven. In COD, targets inherently mimic the background. Using confidence-driven updates leads to confirmation bias, where the model inadvertently amplifies the confidence of erroneous background predictions. Domain adaptation methods (\textit{e.g.,} GH) operate at the domain level, requiring access to target domain datasets, and fail to resolve instance-level feature entanglement. HCL is a sample-specific TTA method that dynamically adapts during inference on a single image. Moreover, HCL is fundamentally consistency-driven. By utilizing hierarchical representation reconstruction and prototype consistency to capture global structures, texture residuals, and robust semantics independently of classification logits, HCL provides an unbiased anchor. 

\textit{3) Analysis of mask strategies.} In Table \ref{tab:mask strategies}, random masking achieves superior performance compared to fixed geometric patterns (round, grid, block). We attribute to: 1) Fixed masking strategies consistently eliminate specific frequency components, creating persistent blind spots in feature space and learning. 2) Random masking dynamically varies the suppressed frequency bands across spatial locations, ensuring no single spectral component is entirely excluded during reconstruction. This encourages the model to holistically integrate multi-spectrum cues rather than over-relying on localized patterns. 3) The stochastic characteristic of random masks enhances robustness against diverse corruption types by preventing overfitting to artificial geometric artifacts.

\textit{4) Analysis of interactive strategies.} In Table \ref{tab:interactive strategy}, our TAG outperforms other cross-attention-based strategies, which can be attributed to: 1) Instead of direct feature-level interaction, we refine internal features along the channel dimension and construct an affinity map, encouraging coherent and compact representations while mitigating noise entanglement; 2) A Top-K selection operator is introduced to filter out low-response values, enabling entropy compression and enhancing the quality of information propagation.

\textit{5) Analysis of prototype integration strategies.} In Table \ref{tab:prototype integration strategy}, variational fusion demonstrates superior performance. We analyze that: 1) Unlike fixed fusion schemes, variational fusion learns distribution-based representations, allowing to dynamically assign confidence-aware weights to different features based on reliability; 2) By explicitly incorporating uncertainty, variational fusion suppresses noisy or redundant information and highlights informative cues, improving robustness in challenging conditions such as low saliency or distribution shifts; 3) Fusion in the probabilistic latent space facilitates smoother integration of complementary cues, enabling more consistent and discriminative representations than simple arithmetic operations or attention that may amplify noise.

\textit{6) Analysis of mask ratios.} In Figure \ref{fig:Hyperparameter settings}, spatial and frequency masking both achieve optimal performance at 25\% mask ratio, as moderate masking balances feature stability and contextual integrity. Excessive spatial masking (\textgreater 25\%) removes critical local details, while excessive frequency masking (\textgreater 25\%) disproportionately disrupts high-frequency textures or low-frequency structures, creating irrecoverable information gaps. Besides, frequency masking shows sharper performance decay due to the entangled spectral components increasing reconstruction ambiguity, whereas spatial masking preserves more structural coherence at higher ratios.

\textit{7) Analysis of Top-K selection operator.} In Figure \ref{fig:Hyperparameter settings}, selecting the top 70\% (80\%) values yields the best performance. When the threshold is lower, performance improves gradually; when it exceeds 70\% (80\%), a performance drop is observed. We argue that: 1) Retaining too few connections (low Top-K ratio) limits the model’s ability to capture sufficient contextual cues, especially long-range dependencies essential for complex scenes; 2) Retaining too many connections introduces noisy or less informative interactions, which may overwhelm the discriminative patterns and degrade the effectiveness of the feature refinement process. The optimal threshold achieves a balance between preserving informative associations and filtering out noise, optimizing both precision and generalization.

\textit{8) Analysis of TTA iterations.} In Figure \ref{fig:Hyperparameter settings}, we observe a turning point at 30 adaptation iterations: performance improves steadily when the iteration count is below 30, and then stabilizes beyond that. We analyze that: 1)During the early testing phase, a clear distribution mismatch exists between the test samples and the model's learned feature space, as indicated by the significant discrepancy between prototypes $\mathbf{P}$ and $\mathbf{P}_{\rm rec}$ distributions; 2) Through iterative refinement (\textit{e.g.,} progressive calibration guided by confidence map $\Phi$), the model gradually aligns its internal parameters with the distribution of samples, reaching a relatively optimal state. Further updates yield diminishing returns, resulting in performance convergence.

\textit{9) Analysis of input size.} In Figure \ref{fig:Hyperparameter settings}, the performance generally exhibits a positive correlation with the input resolution. We attribute this to: 1) Higher-resolution inputs provide denser spectral sampling in the Fourier domain (expanded along the $u$ and $v$ dimensions), enabling the model to more precisely separate critical high-frequency details and structured low-frequency components; 2) The larger pixel space offers greater capacity to preserve discriminative features of the target, mitigating information loss during the encoding process.

\begin{figure*}[tb]
  \centering
  \includegraphics[width=1\textwidth]{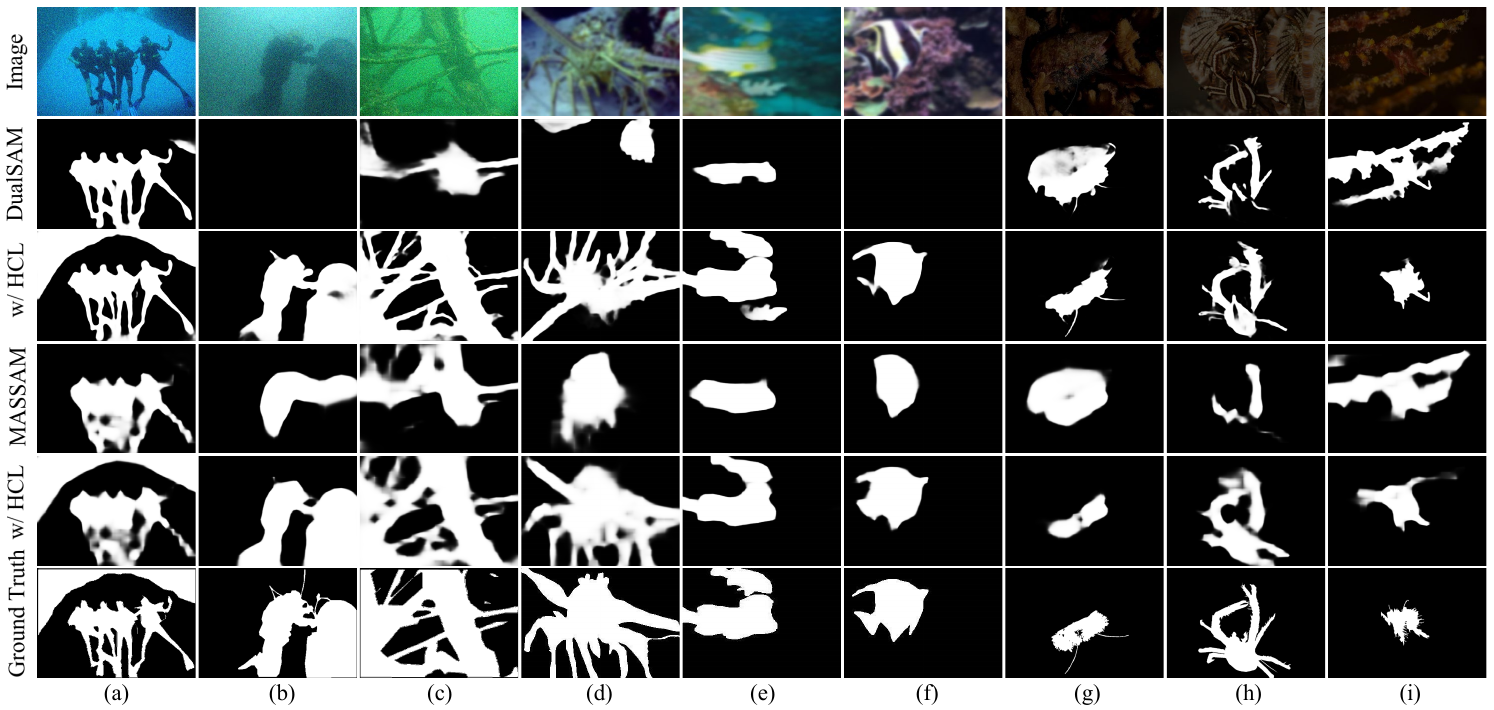}
  \caption{Qualitative comparison on degraded underwater scenarios. (a)-(c): GN setting, (d)-(f): GB setting, (g)-(i): CR setting.}
\label{fig:Underwater_Domain_Vis_Comparision}
\end{figure*}

\begin{figure}[tb]
  \centering
  \includegraphics[width=0.48\textwidth]{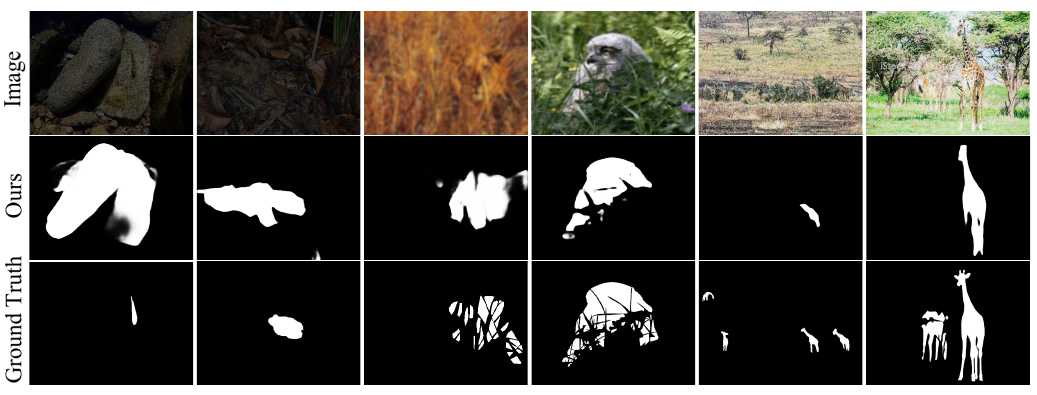}
  \caption{Failure cases under degradation conditions.}
  \label{fig:Failure cases}
\end{figure}

\subsection{Broader Impacts} 
We evaluate state-of-the-art methods on four underwater benchmarks under the same three degradation settings, and further integrate HCL to explore its performance gains.

\textit{1) Quantitative Analysis.} In Table~\ref{tab:Quantitative comparison on degraded underwater benchmark datasets}, we present results under GN, GB, and CR conditions. Across all datasets, GB causes the most severe degradation, followed by GN, while CR shows the least impact. By integrating HCL, both DualSAM and MASSAM consistently achieve performance gains across metrics and datasets. Notably, for DualSAM on the MAS3K, HCL improves $F^{w}_{\beta}$ by 4.2\% and reduces MAE by  1.3\% under GB, highlighting its ability to recover fine details. Under GN and CR, HCL still brings steady gains of 1.2\%–4.4\%

\textit{2) Qualitative Analysis.} In Figure \ref{fig:Underwater_Domain_Vis_Comparision}, we present qualitative comparisons under various challenging scenarios. Integrating HCL boosts comprehensive perception. In multi-object scenes (columns a, b, and e), it preserves complete instance integrity and avoids omissions by leveraging prototype consistency and fusion to suppress interference between adjacent objects. For objects with highly irregular shapes (columns c, d, and h), the variant captures long-range dependencies and preserves boundary details through non-local affinity modeling. In low-saliency settings (column f), where foreground-background contrast is weak, our hierarchical frequency reconstruction enhances subtle discriminative cues, effectively improving foreground-background separation. In camouflaged instances (columns f, g, and h), our method leverages task-guided reconstruction and entropy-based confidence weighting to highlight critical boundaries and suppress false detections.

\textit{3) Why can HCL work under underwater conditions?} 1) HCL leverages spatial- and frequency-decoupled reconstruction for TTA, which is not limited to camouflage scenarios and can be extended to other binary segmentation tasks; 2) Existing underwater benchmarks and learning paradigms also suffer from train-test distribution shifts and fixed model parameters, making TTA essential, especially when degradation further amplifies distributional discrepancies.

\subsection{Limitation and Future Work}
In Figure \ref{fig:Failure cases}. the HCL framework remains challenged in extreme scenarios: highly camouflaged targets (\textit{e.g.,} insects perfectly mimicking leaves) may evade detection as global feature homogenization overshadows subtle discriminative cues, while highly irregular backgrounds (\textit{e.g.,} forest floors with fragmented shadows) disrupt frequency-based analysis, causing localization errors. These limitations reflect the inherent difficulty in balancing structural preservation and noise suppression under environmental uncertainty. We will focus on: 1) Integrating attention with dynamic sparsity to enhance feature disentanglement by focusing on sample-critical frequency bands; 2) Extending the HCL framework to specific scenes (\textit{e.g.,} remote sensing \cite{zha2022multifeature} and agricultural protection \cite{zha2021lightweight}), multimodal learning \cite{zha2025implicit,pei2024emotion,pei2025attentionar}.

\section{CONCLUSION}
This paper tackles the challenge of COD under distribution shifts, where static models struggle to generalize. We propose HCL, a dynamic TTA framework that enables sample-specific self-calibration via spatial-frequency decoupled reconstruction. Key components such as TAG and entropy-guided prototype fusion enhance structural consistency and feature robustness. These findings highlight the importance of dynamic adaptation and suggest HCL’s potential for broader binary segmentation tasks.

\normalem
\bibliographystyle{IEEEtran}
\bibliography{reference_arxiv.bib}

@inproceedings{fan2020pranet,
  title={Pranet: Parallel reverse attention network for polyp segmentation},
  author={Fan, Deng-Ping and Ji, Ge-Peng and Zhou, Tao and Chen, Geng and Fu, Huazhu and Shen, Jianbing and Shao, Ling},
  booktitle={International conference on medical image computing and computer-assisted intervention},
  pages={263--273},
  year={2020},
  organization={Springer}
}

@inproceedings{sun2020test,
  title={Test-time training with self-supervision for generalization under distribution shifts},
  author={Sun, Yu and Wang, Xiaolong and Liu, Zhuang and Miller, John and Efros, Alexei and Hardt, Moritz},
  booktitle={International conference on machine learning},
  pages={9229--9248},
  year={2020},
  organization={PMLR}
}

@inproceedings{niu2022efficient,
  title={Efficient test-time model adaptation without forgetting},
  author={Niu, Shuaicheng and Wu, Jiaxiang and Zhang, Yifan and Chen, Yaofo and Zheng, Shijian and Zhao, Peilin and Tan, Mingkui},
  booktitle={International conference on machine learning},
  pages={16888--16905},
  year={2022},
  organization={PMLR}
}

@article{zha2024dual,
  title={Dual domain perception and progressive refinement for mirror detection},
  author={Zha, Mingfeng and Fu, Feiyang and Pei, Yunqiang and Wang, Guoqing and Li, Tianyu and Tang, Xiongxin and Yang, Yang and Shen, Heng Tao},
  journal={IEEE Transactions on Circuits and Systems for Video Technology},
  volume={34},
  number={11},
  pages={11942--11953},
  year={2024},
  publisher={IEEE}
}

@article{zha2025heterogeneous,
  title={Heterogeneous Experts and Hierarchical Perception for Underwater Salient Object Detection},
  author={Zha, Mingfeng and Wang, Guoqing and Pei, Yunqiang and Li, Tianyu and Tang, Xiongxin and Li, Chongyi and Yang, Yang and Shen, Heng Tao},
  journal={IEEE Transactions on Image Processing},
  year={2025},
  publisher={IEEE}
}

@inproceedings{zha2024weakly,
  title={Weakly-supervised mirror detection via scribble annotations},
  author={Zha, Mingfeng and Pei, Yunqiang and Wang, Guoqing and Li, Tianyu and Yang, Yang and Qian, Wenbin and Shen, Heng Tao},
  booktitle={Proceedings of the AAAI Conference on Artificial Intelligence},
  volume={38},
  number={7},
  pages={6953--6961},
  year={2024}
}

@inproceedings{zhao2023distortion,
  title={Distortion-aware transformer in 360 salient object detection},
  author={Zhao, Yinjie and Zhao, Lichen and Yu, Qian and Sheng, Lu and Zhang, Jing and Xu, Dong},
  booktitle={Proceedings of the 31st ACM International Conference on Multimedia},
  pages={499--508},
  year={2023}
}

@inproceedings{zha_aaai_2026,
  title={Seeing Beyond Illusion: Generalized and Efficient Mirror Detection},
  author={Zha, Mingfeng and Wang, Guoqing and Li, Tianyu and Dong, Wei and Wang, Peng and Yang, Yang},
  booktitle={Proceedings of the AAAI Conference on Artificial Intelligence},
  year={2026}
}

@inproceedings{liu2025language,
  title={Language-Guided Salient Object Ranking},
  author={Liu, Fang and Liu, Yuhao and Xu, Ke and Ye, Shuquan and Hancke, Gerhard Petrus and Lau, Rynson WH},
  booktitle={Proceedings of the Computer Vision and Pattern Recognition Conference},
  pages={29803--29813},
  year={2025}
}

@article{guan2025contrastive,
  title={A Contrastive-Learning Framework for Unsupervised Salient Object Detection},
  author={Guan, Huankang and Lin, Jiaying and Lau, Rynson WH},
  journal={IEEE Transactions on Image Processing},
  year={2025},
  publisher={IEEE}
}

@article{sun2022boundary,
  title={Boundary-guided camouflaged object detection},
  author={Sun, Yujia and Wang, Shuo and Chen, Chenglizhao and Xiang, Tian-Zhu},
  journal={arXiv preprint arXiv:2207.00794},
  year={2022}
}

@inproceedings{zhu2021inferring,
  title={Inferring camouflaged objects by texture-aware interactive guidance network},
  author={Zhu, Jinchao and Zhang, Xiaoyu and Zhang, Shuo and Liu, Junnan},
  booktitle={Proceedings of the AAAI conference on artificial intelligence},
  volume={35},
  number={4},
  pages={3599--3607},
  year={2021}
}

@inproceedings{sun2024frequency,
  title={Frequency-spatial entanglement learning for camouflaged object detection},
  author={Sun, Yanguang and Xu, Chunyan and Yang, Jian and Xuan, Hanyu and Luo, Lei},
  booktitle={European Conference on Computer Vision},
  pages={343--360},
  year={2024},
  organization={Springer}
}

@inproceedings{wu-popNet,
  title={Source-free depth for object pop-out},
  author={Wu, Zongwei and Paudel, Danda Pani and Fan, Deng-Ping and Wang, Jingjing and Wang, Shuo and Demonceaux, C{\'e}dric and Timofte, Radu and Van Gool, Luc},
  booktitle={Proceedings of the IEEE/CVF international conference on computer vision},
  pages={1032--1042},
  year={2023}
}

@article{zhang2023predictive,
  title={Predictive uncertainty estimation for camouflaged object detection},
  author={Zhang, Yi and Zhang, Jing and Hamidouche, Wassim and Deforges, Olivier},
  journal={IEEE Transactions on Image Processing},
  volume={32},
  pages={3580--3591},
  year={2023},
  publisher={IEEE}
}

@inproceedings{zhang2024unlocking,
  title={Unlocking Attributes’ Contribution to Successful Camouflage: A Combined Textual and Visual Analysis Strategy},
  author={Zhang, Hong and Lyu, Yixuan and Yu, Qian and Liu, Hanyang and Ma, Huimin and Yuan, Ding and Yang, Yifan},
  booktitle={European Conference on Computer Vision},
  pages={315--331},
  year={2024},
  organization={Springer}
}

@article{du2025upgen,
  title={UpGen: Unleashing Potential of Foundation Models for Training-Free Camouflage Detection via Generative Models},
  author={Du, Ji and Wu, Jiesheng and Kong, Desheng and Liang, Weiyun and Hao, Fangwei and Xu, Jing and Wang, Bin and Wang, Guiling and Li, Ping},
  journal={IEEE Transactions on Image Processing},
  year={2025},
  publisher={IEEE}
}

@inproceedings{chen2024just,
  title={Just a Hint: Point-Supervised Camouflaged Object Detection},
  author={Chen, Huafeng and Shao, Dian and Guo, Guangqian and Gao, Shan},
  booktitle={European Conference on Computer Vision},
  pages={332--348},
  year={2024},
  organization={Springer}
}

@article{zha2026think,
  title={Think Twice Before Determining: Towards Scene-aware Visual Reasoning for Mirror Detection},
  author={Zha, Mingfeng and Wang, Guoqing and Pei, Yunqiang and Li, Tianyu and Tang, Xiongxin and Ma, Jiayi and Yang, Yang and Shen, Heng Tao},
  journal={IEEE Transactions on Circuits and Systems for Video Technology},
  year={2026},
  publisher={IEEE}
}

@inproceedings{lai2024camoteacher,
  title={CamoTeacher: Dual-Rotation Consistency Learning for Semi-Supervised Camouflaged Object Detection},
  author={Lai, Xunfa and Yang, Zhiyu and Hu, Jie and Zhang, Shengchuan and Cao, Liujuan and Jiang, Guannan and Wang, Zhiyu and Zhang, Songan and Ji, Rongrong},
  booktitle={European Conference on Computer Vision},
  pages={438--455},
  year={2024},
  organization={Springer}
}

@article{li2023zero,
  title={Zero-shot camouflaged object detection},
  author={Li, Haoran and Feng, Chun-Mei and Xu, Yong and Zhou, Tao and Yao, Lina and Chang, Xiaojun},
  journal={IEEE Transactions on Image Processing},
  volume={32},
  pages={5126--5137},
  year={2023},
  publisher={IEEE}
}

@inproceedings{wang2024few,
  title={Few-Shot Camouflaged Object Segmentation},
  author={Wang, Ziqiu and Li, Yuying and Yang, Yang and Li, Yamin and Liu, Gaoyang},
  booktitle={2024 International Joint Conference on Neural Networks (IJCNN)},
  pages={1--10},
  year={2024},
  organization={IEEE}
}

@inproceedings{zhao2024focusdiffuser,
  title={Focusdiffuser: Perceiving local disparities for camouflaged object detection},
  author={Zhao, Jianwei and Li, Xin and Yang, Fan and Zhai, Qiang and Luo, Ao and Jiao, Zicheng and Cheng, Hong},
  booktitle={European Conference on Computer Vision},
  pages={181--198},
  year={2024},
  organization={Springer}
}

@article{chen2024sam,
  title={SAM-COD+: SAM-guided Unified Framework for Weakly-Supervised Camouflaged Object Detection},
  author={Chen, Huafeng and Wei, Pengxu and Guo, Guangqian and Gao, Shan},
  journal={IEEE Transactions on Circuits and Systems for Video Technology},
  year={2024},
  publisher={IEEE}
}

@inproceedings{luo2024vscode,
  title={Vscode: General visual salient and camouflaged object detection with 2d prompt learning},
  author={Luo, Ziyang and Liu, Nian and Zhao, Wangbo and Yang, Xuguang and Zhang, Dingwen and Fan, Deng-Ping and Khan, Fahad and Han, Junwei},
  booktitle={Proceedings of the IEEE/CVF conference on computer vision and pattern recognition},
  pages={17169--17180},
  year={2024}
}

@inproceedings{he2024text,
  title={Text-prompt Camouflaged Instance Segmentation with Graduated Camouflage Learning},
  author={He, Zhentao and Xia, Changqun and Qiao, Shengye and Li, Jia},
  booktitle={Proceedings of the 32nd ACM International Conference on Multimedia},
  pages={5584--5593},
  year={2024}
}

@inproceedings{hui2024endow,
  title={Endow sam with keen eyes: Temporal-spatial prompt learning for video camouflaged object detection},
  author={Hui, Wenjun and Zhu, Zhenfeng and Zheng, Shuai and Zhao, Yao},
  booktitle={Proceedings of the IEEE/CVF Conference on Computer Vision and Pattern Recognition},
  pages={19058--19067},
  year={2024}
}

@article{zhang2025explicit,
  title={Explicit motion handling and interactive prompting for video camouflaged object detection},
  author={Zhang, Xin and Xiao, Tao and Ji, Ge-Peng and Wu, Xuan and Fu, Keren and Zhao, Qijun},
  journal={IEEE Transactions on Image Processing},
  year={2025},
  publisher={IEEE}
}

@article{hao2025simple,
  title={A simple yet effective network based on vision transformer for camouflaged object and salient object detection},
  author={Hao, Chao and Yu, Zitong and Liu, Xin and Xu, Jun and Yue, Huanjing and Yang, Jingyu},
  journal={IEEE Transactions on Image Processing},
  year={2025},
  publisher={IEEE}
}

@inproceedings{pang2024open,
  title={Open-vocabulary camouflaged object segmentation},
  author={Pang, Youwei and Zhao, Xiaoqi and Zuo, Jiaming and Zhang, Lihe and Lu, Huchuan},
  booktitle={European Conference on Computer Vision},
  pages={476--495},
  year={2024},
  organization={Springer}
}

@inproceedings{radford2021learning,
  title={Learning transferable visual models from natural language supervision},
  author={Radford, Alec and Kim, Jong Wook and Hallacy, Chris and Ramesh, Aditya and Goh, Gabriel and Agarwal, Sandhini and Sastry, Girish and Askell, Amanda and Mishkin, Pamela and Clark, Jack and others},
  booktitle={International conference on machine learning},
  pages={8748--8763},
  year={2021},
  organization={PMLR}
}

@article{wang2020tent,
  title={Tent: Fully test-time adaptation by entropy minimization},
  author={Wang, Dequan and Shelhamer, Evan and Liu, Shaoteng and Olshausen, Bruno and Darrell, Trevor},
  journal={arXiv preprint arXiv:2006.10726},
  year={2020}
}

@article{jang2022test,
  title={Test-time adaptation via self-training with nearest neighbor information},
  author={Jang, Minguk and Chung, Sae-Young and Chung, Hye Won},
  journal={arXiv preprint arXiv:2207.10792},
  year={2022}
}

@inproceedings{brahma2023probabilistic,
  title={A probabilistic framework for lifelong test-time adaptation},
  author={Brahma, Dhanajit and Rai, Piyush},
  booktitle={Proceedings of the IEEE/CVF Conference on Computer Vision and Pattern Recognition},
  pages={3582--3591},
  year={2023}
}

@article{zhao2023delta,
  title={Delta: degradation-free fully test-time adaptation},
  author={Zhao, Bowen and Chen, Chen and Xia, Shu-Tao},
  journal={arXiv preprint arXiv:2301.13018},
  year={2023}
}

@inproceedings{ma2024improved,
  title={Improved self-training for test-time adaptation},
  author={Ma, Jing},
  booktitle={Proceedings of the IEEE/CVF Conference on Computer Vision and Pattern Recognition},
  pages={23701--23710},
  year={2024}
}

@inproceedings{liu2024depth,
  title={Depth-aware test-time training for zero-shot video object segmentation},
  author={Liu, Weihuang and Shen, Xi and Li, Haolun and Bi, Xiuli and Liu, Bo and Pun, Chi-Man and Cun, Xiaodong},
  booktitle={Proceedings of the IEEE/CVF Conference on Computer Vision and Pattern Recognition},
  pages={19218--19227},
  year={2024}
}

@inproceedings{hu2024relax,
  title={Relax image-specific prompt requirement in sam: A single generic prompt for segmenting camouflaged objects},
  author={Hu, Jian and Lin, Jiayi and Gong, Shaogang and Cai, Weitong},
  booktitle={Proceedings of the AAAI Conference on Artificial Intelligence},
  volume={38},
  number={11},
  pages={12511--12518},
  year={2024}
}

@inproceedings{jiang2021focal,
  title={Focal frequency loss for image reconstruction and synthesis},
  author={Jiang, Liming and Dai, Bo and Wu, Wayne and Loy, Chen Change},
  booktitle={Proceedings of the IEEE/CVF international conference on computer vision},
  pages={13919--13929},
  year={2021}
}

@inproceedings{pang2022zoom,
  title={Zoom in and out: A mixed-scale triplet network for camouflaged object detection},
  author={Pang, Youwei and Zhao, Xiaoqi and Xiang, Tian-Zhu and Zhang, Lihe and Lu, Huchuan},
  booktitle={Proceedings of the IEEE/CVF Conference on computer vision and pattern recognition},
  pages={2160--2170},
  year={2022}
}

@inproceedings{fan2020camouflaged,
  title={Camouflaged object detection},
  author={Fan, Deng-Ping and Ji, Ge-Peng and Sun, Guolei and Cheng, Ming-Ming and Shen, Jianbing and Shao, Ling},
  booktitle={Proceedings of the IEEE/CVF conference on computer vision and pattern recognition},
  pages={2777--2787},
  year={2020}
}

@inproceedings{zhai2021mutual,
  title={Mutual graph learning for camouflaged object detection},
  author={Zhai, Qiang and Li, Xin and Yang, Fan and Chen, Chenglizhao and Cheng, Hong and Fan, Deng-Ping},
  booktitle={Proceedings of the IEEE/CVF conference on computer vision and pattern recognition},
  pages={12997--13007},
  year={2021}
}

@inproceedings{mei2021camouflaged,
  title={Camouflaged object segmentation with distraction mining},
  author={Mei, Haiyang and Ji, Ge-Peng and Wei, Ziqi and Yang, Xin and Wei, Xiaopeng and Fan, Deng-Ping},
  booktitle={Proceedings of the IEEE/CVF conference on computer vision and pattern recognition},
  pages={8772--8781},
  year={2021}
}

@inproceedings{yang2021uncertainty,
  title={Uncertainty-guided transformer reasoning for camouflaged object detection},
  author={Yang, Fan and Zhai, Qiang and Li, Xin and Huang, Rui and Luo, Ao and Cheng, Hong and Fan, Deng-Ping},
  booktitle={Proceedings of the IEEE/CVF international conference on computer vision},
  pages={4146--4155},
  year={2021}
}

@inproceedings{lv2021simultaneously,
  title={Simultaneously localize, segment and rank the camouflaged objects},
  author={Lv, Yunqiu and Zhang, Jing and Dai, Yuchao and Li, Aixuan and Liu, Bowen and Barnes, Nick and Fan, Deng-Ping},
  booktitle={Proceedings of the IEEE/CVF conference on computer vision and pattern recognition},
  pages={11591--11601},
  year={2021}
}

@article{fan2021concealed,
  title={Concealed object detection},
  author={Fan, Deng-Ping and Ji, Ge-Peng and Cheng, Ming-Ming and Shao, Ling},
  journal={IEEE transactions on pattern analysis and machine intelligence},
  volume={44},
  number={10},
  pages={6024--6042},
  year={2021},
  publisher={IEEE}
}

@inproceedings{zhang2022preynet_mm,
  title={Preynet: Preying on camouflaged objects},
  author={Zhang, Miao and Xu, Shuang and Piao, Yongri and Shi, Dongxiang and Lin, Shusen and Lu, Huchuan},
  booktitle={Proceedings of the 30th ACM international conference on multimedia},
  pages={5323--5332},
  year={2022}
}

@inproceedings{he2023camouflaged,
  title={Camouflaged object detection with feature decomposition and edge reconstruction},
  author={He, Chunming and Li, Kai and Zhang, Yachao and Tang, Longxiang and Zhang, Yulun and Guo, Zhenhua and Li, Xiu},
  booktitle={Proceedings of the IEEE/CVF conference on computer vision and pattern recognition},
  pages={22046--22055},
  year={2023}
}

@article{he2025run,
  title={Run: Reversible unfolding network for concealed object segmentation},
  author={He, Chunming and Zhang, Rihan and Xiao, Fengyang and Fang, Chengyu and Tang, Longxiang and Zhang, Yulun and Kong, Linghe and Fan, Deng-Ping and Li, Kai and Farsiu, Sina},
  journal={arXiv preprint arXiv:2501.18783},
  year={2025}
}

@inproceedings{kirillov2023segment,
  title={Segment anything},
  author={Kirillov, Alexander and Mintun, Eric and Ravi, Nikhila and Mao, Hanzi and Rolland, Chloe and Gustafson, Laura and Xiao, Tete and Whitehead, Spencer and Berg, Alexander C and Lo, Wan-Yen and others},
  booktitle={Proceedings of the IEEE/CVF International Conference on Computer Vision},
  pages={4015--4026},
  year={2023}
}

@inproceedings{hu2023high,
  title={High-resolution iterative feedback network for camouflaged object detection},
  author={Hu, Xiaobin and Wang, Shuo and Qin, Xuebin and Dai, Hang and Ren, Wenqi and Luo, Donghao and Tai, Ying and Shao, Ling},
  booktitle={Proceedings of the AAAI Conference on Artificial Intelligence},
  volume={37},
  number={1},
  pages={881--889},
  year={2023}
}

@inproceedings{cong2023frequency,
  title={Frequency perception network for camouflaged object detection},
  author={Cong, Runmin and Sun, Mengyao and Zhang, Sanyi and Zhou, Xiaofei and Zhang, Wei and Zhao, Yao},
  booktitle={Proceedings of the 31st ACM international conference on multimedia},
  pages={1179--1189},
  year={2023}
}

@inproceedings{huang2023feature,
  title={Feature shrinkage pyramid for camouflaged object detection with transformers},
  author={Huang, Zhou and Dai, Hang and Xiang, Tian-Zhu and Wang, Shuo and Chen, Huai-Xin and Qin, Jie and Xiong, Huan},
  booktitle={Proceedings of the IEEE/CVF conference on computer vision and pattern recognition},
  pages={5557--5566},
  year={2023}
}

@inproceedings{liu2023explicit,
  title={Explicit visual prompting for low-level structure segmentations},
  author={Liu, Weihuang and Shen, Xi and Pun, Chi-Man and Cun, Xiaodong},
  booktitle={Proceedings of the IEEE/CVF Conference on Computer Vision and Pattern Recognition},
  pages={19434--19445},
  year={2023}
}

@article{yin2024camoformer,
  title={Camoformer: Masked separable attention for camouflaged object detection},
  author={Yin, Bowen and Zhang, Xuying and Fan, Deng-Ping and Jiao, Shaohui and Cheng, Ming-Ming and Van Gool, Luc and Hou, Qibin},
  journal={IEEE Transactions on Pattern Analysis and Machine Intelligence},
  year={2024},
  publisher={IEEE}
}

@article{zhao2024spider,
  title={Spider: A Unified Framework for Context-dependent Concept Understanding},
  author={Zhao, Xiaoqi and Pang, Youwei and Ji, Wei and Sheng, Baicheng and Zuo, Jiaming and Zhang, Lihe and Lu, Huchuan},
  journal={arXiv preprint arXiv:2405.01002},
  year={2024}
}

@inproceedings{yu2024exploring,
  title={Exploring Deeper! Segment Anything Model with Depth Perception for Camouflaged Object Detection},
  author={Yu, Zhenni and Zhang, Xiaoqin and Zhao, Li and Bin, Yi and Xiao, Guobao},
  booktitle={Proceedings of the 32nd ACM International Conference on Multimedia},
  pages={4322--4330},
  year={2024}
}

@article{sun2025conditional,
  title={Conditional diffusion models for camouflaged and salient object detection},
  author={Sun, Ke and Chen, Zhongxi and Lin, Xianming and Sun, Xiaoshuai and Liu, Hong and Ji, Rongrong},
  journal={IEEE Transactions on Pattern Analysis and Machine Intelligence},
  volume={47},
  number={4},
  pages={2833--2848},
  year={2025},
  publisher={IEEE}
}

@article{zhang2025comprompter,
  title={COMPrompter: reconceptualized segment anything model with multiprompt network for camouflaged object detection},
  author={Zhang, Xiaoqin and Yu, Zhenni and Zhao, Li and Fan, Deng-Ping and Xiao, Guobao},
  journal={Science China Information Sciences},
  volume={68},
  number={1},
  pages={112104},
  year={2025},
  publisher={Springer}
}

@inproceedings{zhou2025rethinking,
  title={Rethinking Detecting Salient and Camouflaged Objects in Unconstrained Scenes},
  author={Zhou, Zhangjun and Li, Yiping and Zhong, Chunlin and Huang, Jianuo and Pei, Jialun and Li, Hua and Tang, He},
  booktitle={Proceedings of the IEEE/CVF International Conference on Computer Vision},
  pages={22372--22382},
  year={2025}
}

@inproceedings{ren2025multi,
  title={Multi-modal Segment Anything Model for Camouflaged Scene Segmentation},
  author={Ren, Guangyu and Liu, Hengyan and Lazarou, Michalis and Stathaki, Tania},
  booktitle={Proceedings of the IEEE/CVF International Conference on Computer Vision},
  pages={19882--19892},
  year={2025}
}

@inproceedings{yu2025sam,
  title={Sam-ttt: Segment anything model via reverse parameter configuration and test-time training for camouflaged object detection},
  author={Yu, Zhenni and Zhao, Li and Xiao, Guobao and Zhang, Xiaoqin},
  booktitle={Proceedings of the 33rd ACM International Conference on Multimedia},
  pages={4030--4038},
  year={2025}
}

@inproceedings{ranftl2021vision,
  title={Vision transformers for dense prediction},
  author={Ranftl, Ren{\'e} and Bochkovskiy, Alexey and Koltun, Vladlen},
  booktitle={Proceedings of the IEEE/CVF international conference on computer vision},
  pages={12179--12188},
  year={2021}
}

@article{le2019anabranch,
  title={Anabranch network for camouflaged object segmentation},
  author={Le, Trung-Nghia and Nguyen, Tam V and Nie, Zhongliang and Tran, Minh-Triet and Sugimoto, Akihiro},
  journal={Computer vision and image understanding},
  volume={184},
  pages={45--56},
  year={2019},
  publisher={Elsevier}
}

@article{skurowski2018animal,
  title={Animal camouflage analysis: Chameleon database},
  author={Skurowski, Przemys{\l}aw and Abdulameer, Hassan and B{\l}aszczyk, Jakub and Depta, Tomasz and Kornacki, Adam and Kozie{\l}, Przemys{\l}aw},
  journal={Unpublished manuscript},
  volume={2},
  number={6},
  pages={7},
  year={2018}
}

@inproceedings{li2020mas3k,
  title={MAS3K: An open dataset for marine animal segmentation},
  author={Li, Lin and Rigall, Eric and Dong, Junyu and Chen, Geng},
  booktitle={International Symposium on Benchmarking, Measuring and Optimization},
  pages={194--212},
  year={2020},
  organization={Springer}
}

@article{fu2023masnet,
  title={Masnet: A robust deep marine animal segmentation network},
  author={Fu, Zhenqi and Chen, Ruizhe and Huang, Yue and Cheng, En and Ding, Xinghao and Ma, Kai-Kuang},
  journal={IEEE Journal of Oceanic Engineering},
  year={2023},
  publisher={IEEE}
}

@article{islam2020simultaneous,
  title={Simultaneous enhancement and super-resolution of underwater imagery for improved visual perception},
  author={Islam, Md Jahidul and Luo, Peigen and Sattar, Junaed},
  journal={arXiv preprint arXiv:2002.01155},
  year={2020}
}

@article{drews2021underwater,
  title={Underwater image segmentation in the wild using deep learning},
  author={Drews-Jr, Paulo and Souza, Isadora de and Maurell, Igor P and Protas, Eglen V and C. Botelho, Silvia S},
  journal={Journal of the Brazilian Computer Society},
  volume={27},
  pages={1--14},
  year={2021},
  publisher={Springer}
}

@article{hendrycks2019benchmarking,
  title={Benchmarking neural network robustness to common corruptions and perturbations},
  author={Hendrycks, Dan and Dietterich, Thomas},
  journal={arXiv preprint arXiv:1903.12261},
  year={2019}
}

@article{huang2024learning,
  title={Learning to Adapt Using Test-Time Images for Salient Object Detection in Optical Remote Sensing Images},
  author={Huang, Kan and Fang, Leyuan and Tian, Chunwei},
  journal={IEEE Transactions on Geoscience and Remote Sensing},
  year={2024},
  publisher={IEEE}
}

@article{wang2022pvt,
  title={Pvt v2: Improved baselines with pyramid vision transformer},
  author={Wang, Wenhai and Xie, Enze and Li, Xiang and Fan, Deng-Ping and Song, Kaitao and Liang, Ding and Lu, Tong and Luo, Ping and Shao, Ling},
  journal={Computational Visual Media},
  volume={8},
  number={3},
  pages={415--424},
  year={2022},
  publisher={Springer}
}

@inproceedings{he2016deep,
  title={Deep residual learning for image recognition},
  author={He, Kaiming and Zhang, Xiangyu and Ren, Shaoqing and Sun, Jian},
  booktitle={Proceedings of the IEEE conference on computer vision and pattern recognition},
  pages={770--778},
  year={2016}
}

@inproceedings{Kingma2014,
  author = {Kingma, Diederik P. and Welling, Max},
  booktitle = {2nd International Conference on Learning Representations, {ICLR} 2014, Banff, AB, Canada, April 14-16, 2014, Conference Track Proceedings},
  eprint = {http://arxiv.org/abs/1312.6114v10},
  eprintclass = {stat.ML},
  eprinttype = {arXiv},
  title = {{Auto-Encoding Variational Bayes}},
  year = 2014
}

@inproceedings{mirza2022norm,
  title={The norm must go on: Dynamic unsupervised domain adaptation by normalization},
  author={Mirza, M Jehanzeb and Micorek, Jakub and Possegger, Horst and Bischof, Horst},
  booktitle={Proceedings of the IEEE/CVF conference on computer vision and pattern recognition},
  pages={14765--14775},
  year={2022}
}

@article{huang2024gradient,
  title={Gradient harmonization in unsupervised domain adaptation},
  author={Huang, Fuxiang and Song, Suqi and Zhang, Lei},
  journal={IEEE Transactions on Pattern Analysis and Machine Intelligence},
  volume={46},
  number={12},
  pages={10319--10336},
  year={2024},
  publisher={IEEE}
}

@inproceedings{mansour2024ttt,
  title={TTT-MIM: Test-Time Training with Masked Image Modeling for Denoising Distribution Shifts},
  author={Mansour, Youssef and Zhong, Xuyang and Caglar, Serdar and Heckel, Reinhard},
  booktitle={European Conference on Computer Vision},
  pages={341--357},
  year={2024},
  organization={Springer}
}

@inproceedings{zhang2024fantastic,
  title={Fantastic animals and where to find them: Segment any marine animal with dual sam},
  author={Zhang, Pingping and Yan, Tianyu and Liu, Yang and Lu, Huchuan},
  booktitle={Proceedings of the IEEE/CVF Conference on Computer Vision and Pattern Recognition},
  pages={2578--2587},
  year={2024}
}

@article{yan2024mas,
  title={MAS-SAM: Segment Any Marine Animal with Aggregated Features},
  author={Yan, Tianyu and Wan, Zifu and Deng, Xinhao and Zhang, Pingping and Liu, Yang and Lu, Huchuan},
  journal={arXiv preprint arXiv:2404.15700},
  year={2024}
}

@article{zha2022multifeature,
  title={Multifeature transformation and fusion-based ship detection with small targets and complex backgrounds},
  author={Zha, Mingfeng and Qian, Wenbin and Yang, Wenji and Xu, Yilu},
  journal={IEEE Geoscience and Remote Sensing Letters},
  volume={19},
  pages={1--5},
  year={2022},
  publisher={IEEE}
}

@article{zha2021lightweight,
  title={A lightweight YOLOv4-Based forestry pest detection method using coordinate attention and feature fusion},
  author={Zha, Mingfeng and Qian, Wenbin and Yi, Wenlong and Hua, Jing},
  journal={Entropy},
  volume={23},
  number={12},
  pages={1587},
  year={2021},
  publisher={MDPI}
}

@inproceedings{zha2025implicit,
  title={Implicit counterfactual learning for audio-visual segmentation},
  author={Zha, Mingfeng and Li, Tianyu and Wang, Guoqing and Wang, Peng and Wu, Yangyang and Yang, Yang and Shen, Heng Tao},
  booktitle={Proceedings of the IEEE/CVF International Conference on Computer Vision},
  pages={22349--22360},
  year={2025}
}

@inproceedings{pei2024emotion,
  title={Emotion recognition in hmds: A multi-task approach using physiological signals and occluded faces},
  author={Pei, Yunqiang and Tang, Jialei and Tang, Qihang and Zha, Mingfeng and Xie, Dongyu and Wang, Guoqing and Liu, Zhitao and Xie, Ning and Wang, Peng and Yang, Yang and others},
  booktitle={Proceedings of the 32nd ACM International Conference on Multimedia},
  pages={5977--5986},
  year={2024}
}

@inproceedings{pei2025attentionar,
  title={AttentionAR: AR Adaptation and Warning for Real-World Safety via Attention Modeling and MLLM Reasoning},
  author={Pei, Yunqiang and Huang, Renming and Zha, Mingfeng and Wang, Guoqing and Wang, Peng and Kang, Qiao and Yang, Yang and Shen, Heng Tao},
  booktitle={Proceedings of the 38th Annual ACM Symposium on User Interface Software and Technology},
  pages={1--19},
  year={2025}
}

@inproceedings{wang2024depth,
  title={Depth-aware concealed crop detection in dense agricultural scenes},
  author={Wang, Liqiong and Yang, Jinyu and Zhang, Yanfu and Wang, Fangyi and Zheng, Feng},
  booktitle={Proceedings of the IEEE/CVF Conference on Computer Vision and Pattern Recognition},
  pages={17201--17211},
  year={2024}
}

\vfill
\end{document}